%% file: acl_latex.tex
\pdfoutput=1

\documentclass[11pt]{article}

\usepackage[]{acl}

\usepackage{times}
\usepackage{latexsym}
\usepackage[T1]{fontenc}
\usepackage[utf8]{inputenc}
\usepackage{microtype}
\usepackage{inconsolata}

\usepackage{amssymb,mathtools}
\usepackage{algorithm}
\usepackage{algpseudocode}
\usepackage{dsfont}
\usepackage{graphicx,paralist,multirow}
\usepackage{makecell,booktabs}
\usepackage{cleveref}
\crefformat{section}{\S#2#1#3}
\crefformat{subsection}{\S#2#1#3}
\usepackage{xcolor,colortbl}
\usepackage{subcaption}
\usepackage{physics}
\usepackage{comment}

\input{dfn.tex}
\title{Do Localization Methods Actually Localize Memorized Data in LLMs? \\ A Tale of Two Benchmarks} 

\author{
\textbf{Ting-Yun Chang}
\quad\quad \textbf{Jesse Thomason}
\quad\quad \textbf{Robin Jia} \\
University of Southern California, Los Angeles, CA, USA \\
\texttt{\{tingyun, jessetho, robinjia\}@usc.edu}
}

\begin{document}
\maketitle
\begin{abstract}
\input{00_abstract}
\end{abstract}

\input{01_intro}
\input{latex/Figs/mainfig}
\input{02_background}
\input{03_terms}
\input{04_benchmarks}
\input{05_methods}
\input{07_experiments}
\input{08_discussion}
\input{latex/Figs/ig-single-layer}
\input{09_conclusion}
\input{10_limitations}

\input{acknowledgements.tex}
\bibliography{anthology,custom}
\clearpage
\appendix
\input{11_appendix.tex}

\end{document}

%% file: dfn.tex
\newcommand{\IG}{\textsc{IG}}
\newcommand{\Act}{\textsc{Activations}}
\newcommand{\Zero}{\textsc{Zero-Out}}
\newcommand{\Slim}{\textsc{Slimming}}
\newcommand{\HC}{\textsc{Hard Concrete}}
\newcommand{\Rand}{\textsc{Random}}

\newcommand{\BI}{\textsc{INJ} Benchmark}
\newcommand{\BII}{\textsc{DEL} Benchmark}

\newcommand{\MemLoss}{\ell_\theta^\text{mem}}
\newcommand{\Attr}{\mathcal{A}^l(i)}
\newcommand{\Neuron}{{n}^l_i}
\newcommand{\ColVec}{\vb{v}^l_i}
\newcommand{\ActVal}{h^l_i}

\newcommand{\LLM}{\theta}
\newcommand{\LMP}{P_\theta}
\newcommand{\OH}{h^l_t}
\newcommand{\NH}{\hat{h}^l_{t}}
\newcommand{\p}{\operatorname{P}}
\newcommand{\IGP}{\p(\NH)}

\newcommand*\inlineimage[1]{\raisebox{-0.15\baselineskip}{\includegraphics[height=0.98\baselineskip]{#1}$\,$}}
\newcommand{\TP}{\inlineimage{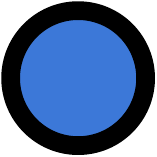}}
\newcommand{\FP}{\inlineimage{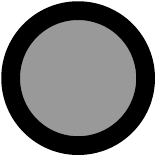}}
\newcommand{\FN}{\inlineimage{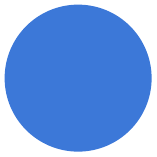}}
\newcommand{\Dropped}{\inlineimage{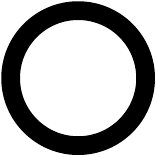}}

\DeclareMathOperator*{\argmax}{argmax}
\newcommand{\Tok}{w}

\newcommand{\Sparse}{\lambda}
\newcommand{\Mask}{m^l_i}
\newcommand{\HCMask}{\hat{m}^l_i}
\newcommand{\ClipMask}{\bar{m}^l}
\newcommand{\Reg}{\mathcal R(\ClipMask)}
\newcommand{\Uniform}{\mathcal{U}\left(0,1\right)}
\newcommand{\T}{\mathcal{T}}

\newcommand{\FTW}{\tilde{\theta_i}}
\newcommand{\N}{N}

%% file: 00_abstract.tex
The concept of localization in LLMs is often mentioned in prior work; however, methods for localization have never been systematically and directly evaluated.
We propose two complementary benchmarks that evaluate the ability of localization methods to pinpoint LLM components responsible for memorized data.
In our \BI, we actively \emph{inject} a piece of new information into a small subset of LLM weights, enabling us to directly evaluate whether localization methods can identify these ``ground truth'' weights. 
In our \BII, we evaluate localization by measuring how much dropping out identified neurons \emph{deletes} a memorized pretrained sequence.
Despite their different perspectives, our two benchmarks yield consistent rankings of five localization methods.
Methods adapted from network pruning perform well on both benchmarks, and all evaluated methods show promising localization ability.
On the other hand, even successful methods identify neurons that are not specific to a single memorized sequence.\footnote{Code link: \url{https://github.com/terarachang/MemPi}}

%% file: 01_intro.tex
\section{Introduction}
\label{sec:intro}
Large language models (LLMs) memorize many sequences from their pretraining corpora~\cite{carlini2019secret, lehman-etal-2021-bert, lee2023language}.
For example, \citet{carlini2021extracting} show that GPT2 \cite{radford2019language} can leak some private contact information verbatim.
This paper studies whether we can \emph{localize} a piece of memorized data, i.e., identify components in LLMs responsible for generating a sequence (near) verbatim.
Successful localization may inform further work in machine unlearning \cite{cao2015towards, bourtoule2021machine}; for instance, one could apply ``neural surgery’’ to the located components to make the LLM forget a piece of sensitive information. 

Prior work on knowledge editing suggests that we can locate a small set of LLM parameters that store factual knowledge~\cite{dai-etal-2022-knowledge,meng2022locating}.
These works demonstrate localization success by showing knowledge editing success when updating only the located LLM parameters.
However, \citet{hase2023does} argue that editing success and localization are actually uncorrelated.
Similarly, prior methods that identify subnetworks in LLMs \cite{gong-etal-2022-finding, panigrahi2023task} usually focus on the performance of downstream classification tasks, lacking direct evaluation on localization per se.
Hence, the degree of existing methods' localization success remains unclear.

This paper studies the open question, ``Do localization methods actually localize memorized data in LLMs?’’ 
We first propose decoupling localization success from downstream success in our \BI.
Our key insight is to actively create the ground-truth weights responsible for data memorization.
Specifically, we force LLMs to use a small set of pre-decided weights to memorize a piece of new information unseen during pretraining.
Therefore, we have the ground-truth locations where the new information is injected.
We can then directly evaluate how well different localization methods recall the indices of the injected weights.

We further apply the localization methods to a real-world scenario: identifying a small set of neurons in an LLM responsible for memorizing a pretrained sequence.
In this setting, evaluating localization success is more challenging because the ground-truth ``location'' of each memorized sequence is unknown.
We propose the \BII, inspired by knockouts \cite{olsson2022context}, a reverse-engineering approach that removes a set of nodes from the computation graph to observe their importance for specific model behavior. 
We first collect a set of memorized sequences, and for each sequence, we drop out the located neurons to measure their importance to memorizing that target sequence.
A successful localization should cleanly erase the target sequence from an LLM without hurting the memorization of the other sequences in the set after dropout.
Our two benchmarks complement each other: the \BI\ provides a direct evaluation of localization methods under a well-controlled setup, while \BII\ answers if the methods can localize pretrained sequences that LLMs have already memorized.

We systematically evaluate five methods on our two benchmarks, including existing localization methods (\Act, \citealp{geva-etal-2022-transformer}; \IG, \citealp{dai-etal-2022-knowledge}), a brute-force method that searches for the most important neurons (\Zero), and two methods we adapt from network pruning \cite{hassibi1992second,han2015deep}, \Slim\ and \HC.
Our two benchmarks rank the five methods in the same order, showing especially strong localization ability for \HC.
For example, dropping out only $0.5\%$ of neurons in Pythia-6.9B \cite{biderman2023pythia} identified by \HC\ makes the model forget $57.7\%$ of the target memorized tokens on average.
On the other hand, the \BII\ shows all methods struggle to balance between erasing the target sequence and retaining other memorized data, indicating that the identified neurons are also relevant for memorizing some other sequences.
Overall, both benchmarks agree all evaluated localization methods are promising, but precise localization of a single sequence remains difficult.

%% file: latex/Figs/mainfig.tex
\begin{figure*}[t!]
  \centering
  \includegraphics[width=0.91\linewidth]{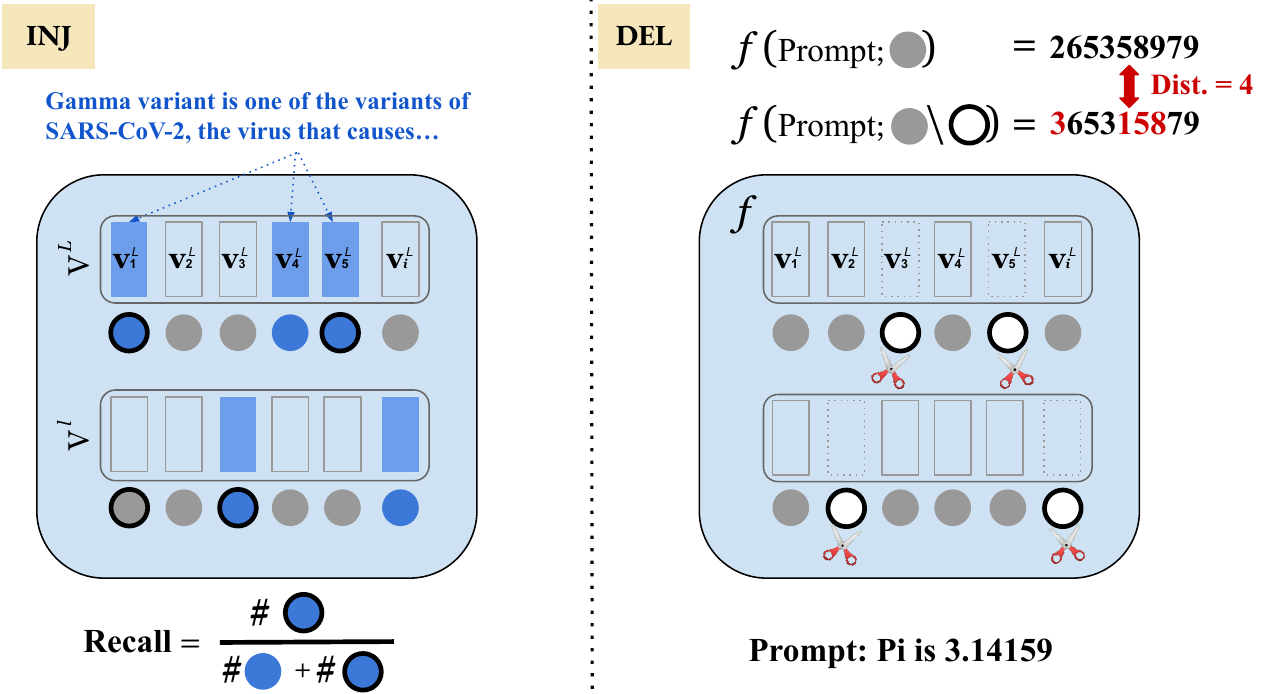}
  \caption{\textbf{Left:} \BI\ updates a small set of LLM weights to store the new piece of data, where the fine-tuned weight vectors and the corresponding neurons are filled with blue. The neurons predicted by a localization method are circled with black. \TP\ denotes true-positive, \FP\ false-positive, and \FN\ false-negative neurons. 
  \textbf{Right:} \BII\ drops out the predicted neurons \Dropped\ on a memorized pretrained sequence.
  A large change in Levenshtein distance after dropout indicates that \Dropped\ were important for LLM $f$ to retrieve the memorized sequence.}
  \label{fig:mainfig}
\end{figure*}

%% file: 02_background.tex
\section{Background and Task Terminology}
\label{sec: ffn-background}

A Transformer layer \cite{vaswani2017attention} consists of multi-head self-attention and a feed-forward network (FFN).
Prior work shows that LLMs use their FFNs rather than self-attention as ``memories’’ to store knowledge \cite{geva-etal-2021-transformer, geva-etal-2022-transformer, meng2022locating}.
Here, an FFN has two fully connected layers with a non-linear activation function $\sigma$:
\begin{align}
    h^l &= \sigma(W^l\,\vb{x}^l) \\
    o^l &= V^l\,h^l, \label{eq: ffn}
\end{align}
where $\vb{x}^l \in \mathbb{R}^{d_1}$ is the input hidden states to the $l$-th FFN layer, $W^l \in \mathbb{R}^{d_2 \times d_1}, V^l \in \mathbb{R}^{d_1 \times d_2}$ are the weights, $h^l \in \mathbb{R}^{d_2}$ the intermediate hidden states, and $o^l \in \mathbb{R}^{d_1}$ the output hidden states.
\citet{geva-etal-2022-transformer} rewrite Eq. \ref{eq: ffn} as a linear combination of columns of $V^l$.
Let $\ColVec \in \mathbb{R}^{d_1}$ be the $i$-th column of $V^l$ and $\ActVal \in \mathbb{R}$ be the $i$-th neuron activation of $h^l \in \mathbb{R}^{d_2}$. We have:
\begin{equation}
    o^l = V^l\,h^l = \sum_{i=1}^{d_2} \ActVal \cdot \ColVec \label{eq: ffn_col}
\end{equation}
They show that different concepts are stored in different $\ColVec$, and that we can view each activation $\ActVal$ as a memory coefficient to retrieve a concept.
 

%% file: 03_terms.tex
\paragraph{Neurons.}
\citet{dai-etal-2022-knowledge} observe the existence of knowledge neurons, a small set of neurons in FFN hidden states $h^l$ that corresponds to a relational fact, where a neuron means a component of the vector $h^l$.
For example, given the input ``\textit{The capital of Ireland is \underline{~~~~}}’’, they can increase the model probability on the correct token ``\textit{Dublin}’’ by amplifying the activation $h^l_i$ of the identified knowledge neurons.
With Eq. \ref{eq: ffn_col}, we can view increasing activation $h^l_i$ as promoting the concept stored in $\ColVec$.

In this work, we only search for neurons in FFNs responsible for memorizing a sequence, following \citet{dai-etal-2022-knowledge}.
In the \BI, we ensure that FFNs act as neural memories by only updating a set of weight vectors $\ColVec$ to memorize the new information.
As each $\ColVec$ corresponds to a neuron in $h^l$, locating the updated weights is equivalent to locating the corresponding neurons.
In the rest of the paper, we refer to neurons as the neurons in $\{h^l\}_{l=1}^L$, where $L$ is the number of layers.

\paragraph{Dropout.}
Different from \citet{srivastava2014dropout}, we drop out located neurons at test time to erase a memorized sequence from the LLM.
We can view dropping out the $i$-th neuron in $h^l$ as excluding the contribution of $\ColVec$ from the output $o^l$ in Eq. \ref{eq: ffn_col}.

\paragraph{Memorized Sequences.} 
Consider a sequence ${x = (p, s)}$ that consists of a prefix $p$ and a suffix $s$.
Given the prefix as the prompt, if an LLM is able to \emph{nearly reconstruct} the suffix with greedy decoding, we say $x$ is a memorized sequence by the LLM.
We discuss in \cref{sec:B2} our criteria on suffix reconstruction, where we tolerate near-verbatim memorization; we also ensure every sequence has a non-trivial suffix.

\paragraph{Localization.} 
\citet{hase2023does} provides a general definition of localization: identifying components of a model responsible for a certain behavior.
Under this definition, we consider components as a small set of neurons and behavior as the LLM's generation of a memorized sequence.
Although some components are necessary for generation, e.g., the input and output token embeddings, we exclude them as they are not specific to a target sequence.

\paragraph{Localization Methods.}
Given an LLM, a memorized sequence $x$, and a fixed number $k$, a localization method outputs $k\%$ of neurons at each layer as the predictions to localize sequence $x$ in the LLM.

%% file: 04_benchmarks.tex
\section{Two Localization Benchmarks}
How do we know whether a method is successful in localization?
We propose two benchmarking approaches: one \emph{injects} a new piece of information into specific parameters in LLMs, while another \emph{deletes} an existing memorized sequence from LLMs via dropout.
A successful localization method should do well on both benchmarks.

\subsection{The \BI}
\label{sec:B1}
A principal challenge in evaluating localization methods is the lack of ground-truth location.
We propose the \BI, which first creates ground truth by actively injecting a piece of unseen information into a small subset of LLM weights.
We can then directly evaluate the correctness of a localization method in predicting the indices of those injected weights.

\paragraph{Data.} The ECBD-2021 dataset \cite{onoe-etal-2022-entity} contains 156 definition sentences of new entities that rose to popularity during the year 2021, e.g., ``\textit{Gamma variant, also known as lineage P.1...}’’.
Since all LLMs we use are trained on corpora released before 2021, the injected weights are the only parameters in the LLMs responsible for memorizing each new definition sequence $x$.

\paragraph{Information Injection.} 
For each new sequence $x_i$ in the dataset, we randomly sample $r\%$ of the weight vectors $\{\vb{v}_1^l, \dotsc, \vb{v}_{d_2}^l\}_{l=1}^L$ across all $L$ layers, and fine-tune them to memorize $x$.
We keep the rest of the model parameters frozen.
To simulate how LLMs learn data during pretraining, we fine-tune with the normal language modeling loss on $x_i$ (Eq. \ref{eq: lm_loss}).
To ensure the entire sequence is well memorized, we keep fine-tuning until we reach a loss $< 0.05$; therefore, we can simply set the first token as the prefix $p$, and the rest of the sequence as the suffix $s$.
Note we sample a different set of weight vectors $\phi_i$ for each sequence $x_i$ and fine-tune a separate model $\FTW$.
Algorithm \ref{algo:inj} shows the exact injection process.

\paragraph{Evaluation.} 
For each model $\FTW$ injected with a sequence $x_i$, a localization method predicts $k\%$ of neurons at each layer and we calculate Recall@$k\%$.
Specifically, given the set of ground-truth neurons corresponding to all the injected weight vectors across layers, $\Gamma_i$, and the set of all predicted neurons, $\hat{\Gamma_i}$, the recall is $\frac{|\Gamma_i \cap \hat{\Gamma_i}|}{|\Gamma_i|}$.
Finally, we average the recall scores across sequences, and thus average over different choices of weights $\phi_i$ for injection.

\input{latex/algo}

\subsection{The \BII}
\label{sec:B2}
The \BII\ studies whether we can localize a naturally memorized sequence after pretraining, which is not answered by the \BI.
We first collect a set of memorized pretrained sequences, and then apply localization methods to identify the responsible neurons for each sequence.
Without ground-truth neurons, we adopt knockouts \cite{li2016understanding, olsson2022context, geva2023dissecting} for evaluation, which measures the importance of model components based on the effect of removing them.
We drop out the located neurons to observe how much they account for memorizing a sequence.
We quantify memorization with two scores: Accuracy and Levenshtein distance.

\paragraph{Accuracy.}
Recall that a sequence $x=(p, s)$ consists of a prefix $p$ and suffix $s$.  
Accuracy calculates the percentage of correct suffix tokens generated by teacher-forcing and argmax decoding. Formally, 
\begin{align}
    \hat{s_t} = & \argmax_{\Tok \in \text{Voc}} \, \LMP (\Tok | p, s_{ < t}), \, t = 1, \dotsc, T \label{eq:argmax} \\
    & \text{Accuracy} = \frac{1}{T} \sum_{t=1}^T \mathds{1} \{\hat{s_t} = s_t \}, \label{eq: acc}
\end{align}
where $T$ denotes the suffix sequence length, $s_t$ the $t$-th true suffix token, $s_{ <t} = [s_1, \dotsc, s_{t-1}]$, $\hat{s_t}$ the $t$-th generated token, $\LMP$ the probability distribution of the LLM parameterized by $\theta$, and Voc the vocabulary.
Higher Accuracy indicates better memorization of the sequence.

\paragraph{Levenshtein distance.}
While Accuracy is defined at a token level, we note tokens often contain several characters, e.g., ``159’’.
For sequences like ``3.14159265’’, every character is important; thus, we also define a memorization score at the character level.
With Eq. \ref{eq:argmax}, we have $\hat{s} = [\hat{s_1}, \dotsc, \hat{s_T}]$.
We calculate Levenshtein distance between the generated suffix $\hat{s}$ and the true suffix $s$.
Lower Levenshtein distance indicates better memorization. 

\paragraph{Data.}
We collect a set of sequences memorized by each LLM, including Pythia-deduped-2.8B, Pythia-deduped-6.9B, and GPT2-XL.
For Pythia models, the pertaining corpus the Pile-dedupe \cite{gao2020pile} is open-sourced, and we use the following criteria to determine which sequences are memorized.
For each candidate sequence $x$, we set the first $32$ tokens as the prefix $p$ to prompt the LLM to reconstruct the suffix $s$ of $48$ tokens.
First, we filter out sequences with Accuracy (Eq. \ref{eq:argmax}, \ref{eq: acc}) lower than $0.9$.
Second, we use greedy decoding to generate the suffix, filtering out those with a Levenshtein distance greater than $20$ characters to the true suffix.
Third, we discard sequences with repetitive tokens (less than 16 distinct tokens in the suffix).
Finally, we deduplicate the remaining sequences based on n-gram Jaccard index.
We obtain 505 memorized sequences for each Pythia model.
For GPT2-XL, we do not have access to its pretraining corpus and find very few memorized sequences from several public corpora with our criteria.
Thus, we actively search for potentially memorized sequences, discovering 105 memorized sequences and categorizing them manually (Table \ref{tab:exampels}).
See \ref{sec:app_data} for details and example sequences.
\input{latex/Tables/examples}

We sample 5 sequences as the dev set to tune the hyperparameters of different methods (see \ref{sec:app_hyper}), using the rest of the collected sequences as the test set.
We quantify the memorization of LLMs on the collected test sets.
Table \ref{tab:before} in the appendix shows that all LLMs have a high average Accuracy ($\sim 100\%$) and a low Levenshtein distance ($\sim 1$ character) to the true suffix, suggesting that the sequences we collect are indeed well memorized.

\paragraph{Evaluation.}
When we evaluate one sequence $x$ in the collected test set $\mathcal{X}$, we consider the rest of the memorized sequences, $\mathcal{X}\setminus \{x\}$, as negative examples.
A successful localization method should make LLMs forget the target sequence (large changes in memorization scores), but still remember the other negative examples (small changes in memorization scores) after dropping out the predicted $k\%$ of neurons at each layer.\footnote{We do not drop out neurons in the bottommost layer, as it hurts LLMs' overall memorization indiscriminately (\cref{sec: main_single}).}
We also calculate the absolute change in perplexity on a batch of $2048$ sequences sampled from the Pile-dedupe, $\mathcal{D}$, to evaluate whether the general language modeling ability remains intact after dropout.

Despite similarities to the evaluation of model editing \cite{sinitsin2020editable, mitchell2021fast}, we can better reflect localization success.
Unlike \citet{meng2022locating} that edit the located weights with gradients, we restrict our operation to neuron dropout.
Because dropout has limited freedom in changing LLMs behavior, successful deletion via dropout requires successful localization; in contrast, gradient-based editing could succeed even without good localization \cite{hase2023does}.

%% file: latex/algo.tex
\begin{algorithm*}[ht]
\small
  \caption{Information Injection}
  \label{algo:inj}
\begin{algorithmic}
\State {\bfseries Input:} The set of new sequences $\mathcal{X}_{\text{ECBD}} = \{x_i\}_{i=1}^{\N}$; pretrained LLM $\theta$ with $L$ layers; injection ratio $r$
\State {\bfseries Output:} The set of fine-tuned LLMs $\mathcal{M} = \{\FTW\}_{i=1}^{\N}$ 

\State Initialize $\mathcal{M} \gets \varnothing$.
\For{$i \gets 1$ to $\N$}
    \State{$\FTW \gets \theta$} \textcolor{gray}{// Initialize from pretrained weights.}
    \State Retrieve all the FFN weight vectors $\Phi_i = \{\vb{v}_1^l, \dotsc, \vb{v}_{d_2}^l\}_{l=1}^L$ from layers $l$ of $\FTW$. 
    \State Set the random seed to $i$.
    \State $\phi_i \gets$ Randomly sample $r\%$ of weight vectors from $\Phi_i$. \textcolor{gray}{// $\phi_i \subset \Phi_i \subset \FTW$}
    \State Fine-tune $\phi_i$ with the language modeling loss on $x_i$ (Eq. \ref{eq: lm_loss}) with remaining weights $\FTW \setminus \phi_i$ frozen.
    \State $\mathcal{M} \gets \mathcal{M} \cup \FTW$.
\EndFor
\State \textbf{return} $\mathcal{M}$
\end{algorithmic}
\end{algorithm*}

%% file: latex/Tables/examples.tex
\begin{table}[!b]
\begin{center}
\resizebox{0.9\columnwidth}{!}
{
\begin{tabular}{llc}
\toprule
\textbf{Category} &  \textbf{Examples} & \textbf{Count}\\
\bottomrule
Quotes & Churchill, Steve Jobs, Trump & 17 \\
Quotes (Book) & Dune, 1984, Bible & 14 \\
Ordered items & Zodiac Signs, US Presidents & 11 \\
Terms of use & MIT License & 10 \\
Poems & The Second Coming & \phantom{1}9 \\
Code & GitHub & \phantom{1}9 \\
Contact Info & A journalist's email & \phantom{1}7 \\
URLs & Reddit, file link & \phantom{1}5 \\
Others & \makecell[l]{long COINBASE ID, meme, \\ Bill of Rights, Pi digits}
 & 23\\

\bottomrule
\end{tabular}
}
\end{center}
\caption{Collected sequences memorized by GPT2-XL.}
\label{tab:exampels}
\end{table}

%% file: 05_methods.tex
\section{Localization Methods}
\label{sec:method}
We benchmark five localization methods.
Each method assigns an attribution score $\Attr$ to each neuron $\Neuron$, the $i$-th neuron in the $l$-th layer, representing its importance in memorizing a sequence $x$.
At test time, we select the top-$k\%$ of neurons in each layer for each method in terms of attribution scores as the located neurons for $x$ by that method.

Several methods involve calculating the language modeling loss of an LLM $\LLM$ on the suffix of the target sequence $x  = (p, s)$.
We denote the loss as \emph{memorization loss}, $\MemLoss(x)$. Formally,
\begin{equation}
    \MemLoss(x) = \frac{1}{T}\sum_{t=1}^{T} - \log \LMP (s_t | p, s_{ < t}) \label{eq: mem_loss}
\end{equation}

\paragraph{\Zero.}
We introduce an exhaustive method that drops out neurons one by one and uses the resulting change in memorization loss on $x$ as the attribution score to each neuron $\Neuron$: 
\begin{equation}
    \Attr = \ell_{\theta \setminus \Neuron}^\text{mem}(x) - \MemLoss(x)
\end{equation}
We denote $\ell^\text{mem}_{\theta \setminus \Neuron}$ as the memorization loss of the LLM $\LLM$ after dropping out a neuron $\Neuron$.
The larger the change in the loss, the more important the neuron is for memorization.
\Zero\ is closely related to the occlusion-based attribution method \cite{zeiler2014visualizing}.

\paragraph{\Act.}
We can view the neuron activation $h^l_i$ as the memory coefficients (\cref{sec: ffn-background}).
Thus, similar to \citet{geva-etal-2022-transformer}, we set the attribution $\Attr$ as the absolute value of $h^l_i$ multiplied by the vector norm of $\ColVec$, averaged across the suffix length $T$:
\begin{equation}
    \Attr = \frac{1}{T} \sum_{t=1}^T |h^l_{i, t}| \, \|\ColVec\|, \label{eq:act}
\end{equation}
where $h^l_{i, t}$ denotes the activation value at the $t$-th timestep, when the input consists of all the tokens before $s_t$, i.e., $[p, s_{ <t}]$.

\paragraph{Integrated Gradients (\IG).}
We benchmark integrated gradients \cite{sundararajan2017axiomatic}, an attribution method that has been used to identify knowledge neurons and privacy neurons \cite{dai-etal-2022-knowledge, wu-etal-2023-depn}.
\IG\ cumulates the gradients at all points along the path from a zero vector to the original hidden state $h^l$.
See \ref{sec:app_ig} for more details.

\input{latex/Tables/ECBD}
\paragraph{\Slim.} 
We introduce \Slim, a localization method adapted from prior work \cite{liu2017learning, chen-etal-2021-earlybert} on network pruning.
Pruning aims to reduce the model size by finding a subnetwork that can achieve a low loss on the task, e.g., sentiment analysis.
In our setting, we find a small set of neurons that are crucial for maintaining a low memorization loss $\MemLoss(x)$ on \emph{one} target sequence $x$ (Eq. \ref{eq: mem_loss}).
Specifically, \Slim\ minimizes the memorization loss while learning a sparse mask $m^l \in \mathbb{R}^{d_2}$ on the hidden state $h^l$ in every layer, with mask value $\Mask$ on neuron $\Neuron$.
At each layer, we transform $h^l$ to $h^l \odot m^l$ before computing further layers, where $\odot$  denotes element-wise multiplication.
The sparse mask encourages the LLM to use only a small set of neurons to recall a piece of memory.
All the weights of the LLM are kept frozen during the training; only the mask $m^l$ is learnable. Formally,
\begin{align}
    \min_{\substack{m^l\\l = 1, \dotsc, L}} \; \MemLoss(x) + \Sparse \sum_{l=1}^L \|m^l \|_{1},
\end{align}
where $\Sparse$ is the hyperparameter to balance the memorization loss and the $L_1$ sparsity regularization on the mask.
After training, we set the attribution score $\Attr = \Mask$, as $\Mask$ learns the importance of the existence of a neuron to the memorization loss.

\paragraph{\HC.} The limitation of \Slim\ is that it tends to assign mask values $\Mask$ between $0$ and $1$ on most neurons, creating a mismatch between training and testing. 
In particular, at inference time we either activate (equivalent to setting $\Mask = 1$) or drop out ($\Mask = 0$) a neuron.
Thus, we adapt another pruning method \HC\ \cite{louizos2017learning, zheng-etal-2022-robust} for localization, which improves over \Slim\ by encouraging mask values $\Mask$ to be approximately binary.
Similar to \Slim, \HC\ learns parameters $m^l \in \mathbb{R}^{d_2}$ in every layer.
But instead of directly using $m^l$ as the mask, the mask $\ClipMask$ in \HC\ is a random variable (r.v.) that depends on $m^l$.
Specifically, \HC\ derives the mask value $\ClipMask_i$ from a binary concrete \cite{maddison2016concrete,jang2016categorical} random variable, $\HCMask$.
A binary concrete distribution $\HCMask \sim \text{Concrete}(\Mask, \beta)$ is parameterized by the location $\Mask$ and temperature $\beta$.
When the hyperparameter $\beta \rightarrow 0$, sampling from the binary concrete distribution is identical to sampling from a Bernoulli distribution but loses the differentiable property.
With $\beta > 0$, we allow gradient-based optimization of parameter $\Mask$ through the reparametrization trick.
Formally,
\begin{align}
    u_i & \sim \Uniform, \\
    \HCMask &= 
        \sigma \left(\frac{1}{\beta}(\log \frac{u_i}{1-u_i} + \log \Mask) \right), \label{eq:bin_rv}
\end{align}
where $\sigma$ denotes the sigmoid function and $u_i$ is a r.v. sampled from uniform distribution $\Uniform$.
We describe the details about how \citet{louizos2017learning} extend a hard concrete r.v. $\ClipMask$ from the binary concrete r.v. $\HCMask$ and use $L_0$ regularization $\Reg$ to encourage sparsity in \ref{sec:app_HC}. 

To learn the parameters $m^l$, we freeze the LLM weights $\theta$ and simultaneously optimize the memorization loss on the target sequence $x$ and the sparsity loss $\Reg$. Formally,
\begin{align}
    \min_{\substack{m^l\ \\l = 1, \dotsc, L}} \; \MemLoss(x) + \Sparse \sum_{l=1}^L \Reg 
\end{align}
At test time, $\HCMask$ can be estimated as $\sigma \left(\log \Mask \right)$ \cite{louizos2017learning}; thus, we set the attribution score $\Attr = \sigma \left(\log \Mask \right)$.

%% file: latex/Tables/ECBD.tex
\begin{table*}[th]
\begin{center}
\centering
\resizebox{2.\columnwidth}{!}
{
\begin{tabular}{lcccccccccccc}
\toprule
      & \multicolumn{3}{c}{\textbf{GPT2 124M}} 
      & \multicolumn{3}{c}{\textbf{GPT2-XL 1.5B}} 
      & \multicolumn{3}{c}{\textbf{Pythia-deduped 2.8B}} 
      & \multicolumn{3}{c}{\textbf{Pythia-deduped 6.9B}}\\

\cmidrule(lr){2-4}
\cmidrule(lr){5-7}
\cmidrule(lr){8-10}
\cmidrule(lr){11-13}

& R@1\% & R@2\% & R@5\% & R@1\% & R@2\% & R@5\% & R@1\% & R@2\% & R@5\% & R@1\% & R@2\% & R@5\%\\
\midrule
\emph{ratio = 1\%}\\
~  \HC & $\textbf{49.5}_{\hspace{0.05cm}0.48}$ & $\textbf{70.2}_{\hspace{0.05cm}0.54}$ & $\textbf{87.4}_{\hspace{0.05cm}0.33}$ & $\textbf{29.7}_{\hspace{0.05cm}0.47}$ & $\textbf{37.1}_{\hspace{0.05cm}0.49}$ & $\textbf{48.1}_{\hspace{0.05cm}0.46}$ & $34.3_{\hspace{0.05cm}0.33}$ & $50.1_{\hspace{0.05cm}0.43}$ & $\textbf{72.1}_{\hspace{0.05cm}0.47}$ & $36.8_{\hspace{0.05cm}0.45}$ & $\textbf{55.1}_{\hspace{0.05cm}0.51}$ & $\textbf{76.4}_{\hspace{0.05cm}0.41}$ \\
~  \Slim & $48.1_{\hspace{0.05cm}0.64}$ & $66.7_{\hspace{0.05cm}0.69}$ & $80.7_{\hspace{0.05cm}0.54}$ & $19.3_{\hspace{0.05cm}0.49}$ & $29.2_{\hspace{0.05cm}0.59}$ & $41.1_{\hspace{0.05cm}0.59}$ & $\textbf{37.0}_{\hspace{0.05cm}0.43}$ & $\textbf{50.7}_{\hspace{0.05cm}0.47}$ & $61.5_{\hspace{0.05cm}0.44}$ & $\textbf{39.9}_{\hspace{0.05cm}0.38}$ & $\textbf{55.1}_{\hspace{0.05cm}0.38}$ & $66.5_{\hspace{0.05cm}0.35}$ \\
~  \Zero & $24.9_{\hspace{0.05cm}0.78}$ & $37.5_{\hspace{0.05cm}1.05}$ & $53.8_{\hspace{0.05cm}1.24}$ & \phantom{1}$4.1_{\hspace{0.05cm}0.13}$ & \phantom{1}$7.2_{\hspace{0.05cm}0.23}$ & $13.7_{\hspace{0.05cm}0.42}$ & $10.6_{\hspace{0.05cm}0.20}$ & $15.0_{\hspace{0.05cm}0.24}$ & $21.4_{\hspace{0.05cm}0.30}$ & - & - & - \\
~  \IG & $20.5_{\hspace{0.05cm}0.55}$ & $32.1_{\hspace{0.05cm}0.80}$ & $49.9_{\hspace{0.05cm}0.99}$ & \phantom{1}$4.3_{\hspace{0.05cm}0.13}$ & \phantom{1}$7.2_{\hspace{0.05cm}0.21}$ & $13.3_{\hspace{0.05cm}0.37}$ & $11.6_{\hspace{0.05cm}0.22}$ & $16.9_{\hspace{0.05cm}0.28}$ & $23.9_{\hspace{0.05cm}0.34}$ & $12.8_{\hspace{0.05cm}0.23}$ & $18.7_{\hspace{0.05cm}0.29}$ & $27.2_{\hspace{0.05cm}0.35}$ \\
~  \Act & \phantom{1}$3.0_{\hspace{0.05cm}0.09}$ & \phantom{1}$5.2_{\hspace{0.05cm}0.13}$ & $13.3_{\hspace{0.05cm}0.32}$ & \phantom{1}$2.1_{\hspace{0.05cm}0.05}$ & \phantom{1}$5.0_{\hspace{0.05cm}0.10}$ & $12.0_{\hspace{0.05cm}0.16}$ & \phantom{1}$7.8_{\hspace{0.05cm}0.11}$ & $12.8_{\hspace{0.05cm}0.20}$ & $30.5_{\hspace{0.05cm}0.53}$ & \phantom{1}$7.9_{\hspace{0.05cm}0.11}$ & $12.4_{\hspace{0.05cm}0.17}$ & $27.3_{\hspace{0.05cm}0.43}$ \\
~  \Rand & \phantom{1}$1.0_{\hspace{0.05cm}0.04}$ & \phantom{1}$2.1_{\hspace{0.05cm}0.06}$ & \phantom{1}$5.0_{\hspace{0.05cm}0.09}$ & \phantom{1}$1.0_{\hspace{0.05cm}0.01}$ & \phantom{1}$2.0_{\hspace{0.05cm}0.02}$ & \phantom{1}$5.0_{\hspace{0.05cm}0.03}$ & \phantom{1}$1.0_{\hspace{0.05cm}0.01}$ & \phantom{1}$2.0_{\hspace{0.05cm}0.02}$ & \phantom{1}$5.0_{\hspace{0.05cm}0.03}$ & \phantom{1}$1.0_{\hspace{0.05cm}0.01}$ & \phantom{1}$2.0_{\hspace{0.05cm}0.02}$ & \phantom{1}$5.0_{\hspace{0.05cm}0.02}$ \\

\midrule

\emph{ratio = 0.1\%} & @0.1\% & @0.2\% & @0.5\% & @0.1\% & @0.2\% & @0.5\% & @0.1\% & @0.2\% & @0.5\% & @0.1\% & @0.2\% & @0.5\%\\
\cmidrule(lr){2-4}
\cmidrule(lr){5-7}
\cmidrule(lr){8-10}
\cmidrule(lr){11-13}
~  \HC & $56.4_{\hspace{0.05cm}0.83}$ & $79.6_{\hspace{0.05cm}0.89}$ & $93.7_{\hspace{0.05cm}0.52}$ & $\textbf{47.5}_{\hspace{0.05cm}0.40}$ & $\textbf{59.1}_{\hspace{0.05cm}0.47}$ & $68.0_{\hspace{0.05cm}0.46}$ & $\textbf{48.5}_{\hspace{0.05cm}0.49}$ & $\textbf{67.3}_{\hspace{0.05cm}0.50}$ & $\textbf{86.7}_{\hspace{0.05cm}0.34}$ & $46.4_{\hspace{0.05cm}0.60}$ & $\textbf{66.3}_{\hspace{0.05cm}0.71}$ & $\textbf{82.3}_{\hspace{0.05cm}0.48}$ \\
~  \Slim & $\textbf{58.9}_{\hspace{0.05cm}0.59}$ & $\textbf{83.5}_{\hspace{0.05cm}0.68}$ & $\textbf{94.4}_{\hspace{0.05cm}0.49}$ & $35.4_{\hspace{0.05cm}0.56}$ & $55.9_{\hspace{0.05cm}0.64}$ & $\textbf{69.5}_{\hspace{0.05cm}0.55}$ & $48.3_{\hspace{0.05cm}0.43}$ & $63.5_{\hspace{0.05cm}0.46}$ & $73.9_{\hspace{0.05cm}0.43}$ & $\textbf{48.5}_{\hspace{0.05cm}0.57}$ & $60.9_{\hspace{0.05cm}0.60}$ & $71.0_{\hspace{0.05cm}0.71}$ \\
~  \Zero & $54.1_{\hspace{0.05cm}0.68}$ & $77.8_{\hspace{0.05cm}0.78}$ & $90.9_{\hspace{0.05cm}0.70}$ & $14.3_{\hspace{0.05cm}0.62}$ & $21.8_{\hspace{0.05cm}0.94}$ & $31.9_{\hspace{0.05cm}1.27}$ & $16.5_{\hspace{0.05cm}0.48}$ & $21.1_{\hspace{0.05cm}0.57}$ & $26.6_{\hspace{0.05cm}0.66}$ & - & - & - \\
~  \IG & $53.5_{\hspace{0.05cm}0.78}$ & $74.1_{\hspace{0.05cm}0.92}$ & $84.8_{\hspace{0.05cm}0.80}$ & $13.8_{\hspace{0.05cm}0.53}$ & $20.3_{\hspace{0.05cm}0.79}$ & $29.7_{\hspace{0.05cm}1.06}$ & $18.0_{\hspace{0.05cm}0.49}$ & $23.3_{\hspace{0.05cm}0.60}$ & $30.2_{\hspace{0.05cm}0.68}$ & $29.3_{\hspace{0.05cm}1.03}$ & $34.4_{\hspace{0.05cm}1.02}$ & $39.6_{\hspace{0.05cm}0.97}$ \\
~  \Act & $11.1_{\hspace{0.05cm}0.43}$ & $26.5_{\hspace{0.05cm}0.84}$ & $51.5_{\hspace{0.05cm}1.06}$ & \phantom{1}$7.5_{\hspace{0.05cm}0.35}$ & $15.9_{\hspace{0.05cm}0.61}$ & $30.6_{\hspace{0.05cm}0.76}$ & $21.6_{\hspace{0.05cm}0.72}$ & $34.6_{\hspace{0.05cm}0.98}$ & $52.5_{\hspace{0.05cm}1.07}$ & $34.0_{\hspace{0.05cm}1.03}$ & $45.9_{\hspace{0.05cm}1.02}$ & $59.5_{\hspace{0.05cm}0.97}$ \\
~  \Rand & \phantom{1}$0.1_{\hspace{0.05cm}0.03}$ & \phantom{1}$0.2_{\hspace{0.05cm}0.06}$ & \phantom{1}$0.5_{\hspace{0.05cm}0.07}$ & \phantom{1}$0.1_{\hspace{0.05cm}0.01}$ & \phantom{1}$0.2_{\hspace{0.05cm}0.02}$ & \phantom{1}$0.5_{\hspace{0.05cm}0.03}$ & \phantom{1}$0.1_{\hspace{0.05cm}0.01}$ & \phantom{1}$0.2_{\hspace{0.05cm}0.02}$ & \phantom{1}$0.5_{\hspace{0.05cm}0.03}$ & \phantom{1}$0.1_{\hspace{0.05cm}0.01}$ & \phantom{1}$0.2_{\hspace{0.05cm}0.02}$ & \phantom{1}$0.5_{\hspace{0.05cm}0.02}$ \\

\bottomrule
\end{tabular}}

\end{center}

\caption{The \BI. We experiment with injection ratio at $1\%$ (\textbf{Top}) and $0.1\%$ (\textbf{Bottom}) and report the Recall@$k\%$ and standard errors of different localization methods averaged across the sequences in ECBD-2021.}
\label{table:ecbd}
\end{table*}

%% file: 07_experiments.tex
\section{Experiments}

\input{latex/Tables/Pile}
\subsection{\BI\ Results}
\label{sec:b1_res}
Table~\ref{table:ecbd} shows the average Recall@$k\%$ and standard errors of different localization methods on four LLMs under our \BI\ evaluation.
When the injection ratio is $1\%$ (Table~\ref{table:ecbd}; Top), there are $1\%$ of weight vectors injected with each new sequence, yielding $1\%$ of ground truth neurons, and every method predicts $k=\{1, 2, 5\}\%$ of neurons at each layer.
When the injection ratio is $0.1\%$ (Table~\ref{table:ecbd}; Bottom), every method predicts $\{0.1, 0.2, 0.5\}\%$ of neurons at each layer.
We also study the alternative that predicts top-$k$ neurons \emph{across} layers in \ref{sec:app_global}, which shows results consistent with Table~\ref{table:ecbd} but with lower recall overall.

\paragraph{All methods can do localization.}
First, all five localization methods greatly outperform \Rand, which randomly predicts the same number of neurons at each layer.
Interestingly, when the injection ratio is lower ($0.1\%$), all localization methods achieve higher recall, possibly because the information is more concentrated in the injected weights and thus easier to identify.

\paragraph{Pruning-based methods perform the best.}
\Slim\ and \HC, the methods based on network pruning, substantially outperform the other methods across all setups.
Specifically, \HC\ achieves Recall@$0.5\%$ higher than $80$ in three out of four LLMs.
\Zero\ and \IG\ perform similarly and outperform the simple method \Act\ overall, but are much more computationally expensive than the other methods (see comparisons in \ref{sec:app_cost}).

\paragraph{Our results hold under more data and different random seeds.}
In the appendix, we show that our conclusions hold when expanding the \BI\ to the newly released ECBD 2022, 2023 dataset \cite{padmanabhan2024propagating} (\ref{sec:app_more_data}), and they are robust to the choice of random seed, which controls the choice of injected weights (\ref{sec:app_more_seeds}).

\subsection{\BII\ Results}
\label{sec:b2_res}
Table~\ref{table:pile} shows to what extent dropping out ${k=\{0.1, 0.5\}\%}$ of neurons predicted by different methods makes LLMs forget the target sequence $x$  (\textcolor{blue}{Self}), while still memorizing the other sequences $\mathcal{X}\setminus \{x\}$ (\textcolor{purple}{Neg}), and keeping the perplexity on the random batch $\mathcal{D}$ (\textcolor{purple}{Rand-PPL}) intact.
We evaluate one target sequence at a time and report the average absolute changes ($\Delta$) in Accuracy (Acc), Levenshtein distance (Dist), and perplexity after dropout.

\paragraph{All methods show evidence of localization.}
Randomly dropping out the same number of neurons (\Rand) barely changes the LLM behavior.
In comparison, all five localization methods successfully identify neurons that contribute much more to memorizing the target sequence than to negative examples, showing evidence of their localization ability on real-world memorized data.

\paragraph{Methods trade off between \textcolor{blue}{$\Delta$ Self} and \textcolor{purple}{$\Delta$ Neg}.}
We find \Slim\ and \HC\ much more effective than other methods in erasing the target sequence itself.
However, they are worse at preserving LLM memorization of the negative examples and the perplexity of randomly sampled sequences.
For example, dropping out $0.5\%$ of GPT2 neurons predicted by \Slim\ decreases Accuracy by $57.8\%$ and increases $75.4$ characters in Levenshtein distance on the target sequence, but it also hurts the Accuracy on negative examples by $6.4\%$ and increases Levenshtein distance by $7.5$ on average.
On the other hand, \IG\ best maintains the performance on negative examples and perplexity, but is not as successful in erasing the target sequence itself.
Interestingly, although \Zero\ assigns the attribution scores with a leave-one-out approach, it does not perform the best on either target sequences or negative examples, implying that the individual neuron dropout effect does not reliably predict the collective effect of dropping out many neurons at the same time.
Overall, it is challenging for methods to effectively and specifically locate the target sequence at the same time.

\paragraph{Which negative examples are forgotten?}
We analyze how the negative examples affected by dropout are related to the target sequence.
Figure \ref{fig:heatmap} is the confusion matrix on a representative subset of GPT2 memorized data, $\mathcal{Y} \subset \mathcal{X}$, where each row shows how dropping out $0.5\%$ of the neurons predicted by \HC\ on a target sequence changes the Accuracy of every sequence in $\mathcal{Y}$.
We group sequences under the same category (see Table \ref{tab:exampels}) in adjacent rows.
We find \HC\ sometimes confuses related data; for example, in the Address category consisting of mailing addresses, dropping out the neurons of an address sequence also causes substantial Accuracy drops on other addresses.
We also find confusion across the Poems, Shakespeare, and Bible categories of literary sequences.
Qualitatively, we found several web pages containing famous quotes from different poems and books; such co-occurrences may also appear multiple times in GPT2's pretraining corpus and may explain why in Figure \ref{fig:heatmap}, a small set of neurons affect quotes from different sources.
While these findings could suggest that \HC\ struggles to pinpoint neurons that are specific to a target sequence, it may also be that LLMs actually use a shared set of neurons to memorize several related sequences.
Figure \ref{fig:app_heatmap_all} in \ref{sec:app_data} shows the confusion matrices of other methods and Figure \ref{fig:app_heatmap_hc} is the matrix of the entire dataset $\mathcal{X}$.
Both figures share patterns similar to Figure \ref{fig:heatmap}.

\input{latex/Figs/heatmap}
\subsection{Concurrence of the two benchmarks}
This section studies if the two benchmarks rank the methods similarly \cite{liu-etal-2023-question} and whether the differences between methods are significant.
\paragraph{Rankings of localization methods.} The \BI, which solely evaluates the injected target sequences,\footnote{\BI\ does not have negative examples, since we do not have ground-truth neurons of pretrained sequences.} and the \textcolor{blue}{Self-} part of the \BII\ show consistent rankings: \HC\ performs slightly better than \Slim, followed by \Zero\ and \IG; \Act\ performs the worst but still substantially outperforms \Rand.
This consistency suggests that despite the differences in data and setups, the two benchmarks reflect the same underlying localization abilities of different methods.
We believe the reason pruning-based methods perform better is that they learn to mask multiple neurons simultaneously, while other methods only consider the importance of each neuron individually.

\paragraph{Tests of significance.}
We run t-tests to test if pruning-based methods outperform \IG\ significantly.
For the \BI, each method has 24 Recall@$k\%$ scores in Table \ref{table:ecbd}; we run 24 one-tailed paired t-tests accordingly.
With Bonferroni correction, we set the significance level $\alpha = \frac{0.05}{24}$.
Table \ref{table:app_pvalues_inj} in the appendix shows that for \HC\ vs. \IG\ and \Slim\ vs. \IG, respectively, there are 23/24 and 24/24 tests that have p-values $< \alpha$.
Similarly, in the \BII, each method has 6 \textcolor{blue}{$ \Delta$ Self-Acc} scores in Table \ref{table:pile}; thus, we run 6 paired t-tests. 
Table \ref{table:app_pvalues_del} shows that 5/6 and 6/6 tests have p-values $< \frac{0.05}{6}$, for \Slim\ vs. \IG\ and \HC\ vs. \IG, respectively.
Notably, for both benchmarks, most tests have p-values $< 10^{-10}$. 
Overall, these results support our claims that the difference between pruning-based methods and \IG\ is significant.

\subsection{Is the memory of a piece of information distributed over layers?}
\label{sec: main_single}
To understand the individual effect of each layer on memorization, we study the alternative that drops out the same number of neurons in a \emph{single} layer.
In \cref{sec:b2_res}, a method predicts top-$0.1\%$ of neurons in \emph{every} layer after the bottommost layer; thus, we have a ``budget’’ of $N = 0.1\% \times 6400 \times (48-1)$ neurons for GPT2-XL.
Here, the alternative strategy drops out the top-$N$ neurons in a single layer in terms of the attribution scores assigned by a method.

Using the attribution assigned by \IG, Figure \ref{fig:ig-single} compares the two dropout strategies, illustrating their \textcolor{blue}{$\Delta$ Self-Acc} and \textcolor{purple}{$\Delta$ Neg-Acc} scores (see more methods in \ref{sec:single}).
First, we find dropping out neurons in multiple layers much more efficient in erasing the target sequence, as the horizontal blue line shows a greater decrease in \textcolor{blue}{Self-Acc} than the solid blue line, suggesting that memorized information is stored in a distributed fashion over many layers, not concentrated in a single layer.
The only exception is dropping out neurons in Layer 1; however, it also greatly hurts \textcolor{purple}{Neg-Acc}.
The large memorization decreases on all sequences may imply that the bottom layers of LLMs mainly work on processing basic and general information \cite{tenney-etal-2019-bert}, instead of focusing on a specific sequence.

%% file: latex/Tables/Pile.tex
\begin{table*}[!t]
\begin{center}
\centering
\resizebox{1.84\columnwidth}{!}
{
\begin{tabular}{lcccccccccc}
\toprule
      & \multicolumn{2}{c}{ \textcolor{blue}{ $\Delta$ \textbf{Self-Acc} $\downarrow$}} 
      & \multicolumn{2}{c}{\textcolor{blue}{\textbf{ $\Delta$ Self-Dist} $\uparrow$}}
      & \multicolumn{2}{c}{\textcolor{purple}{ $\Delta$ \textbf{Neg-Acc} $\uparrow$}}
      & \multicolumn{2}{c}{\textcolor{purple}{$\Delta$ \textbf{Neg-Dist} $\downarrow$}}
      & \multicolumn{2}{c}{\textcolor{purple}{$\Delta$ \textbf{Rand-PPL} $\downarrow$}}
      \\

\cmidrule(lr){2-3}
\cmidrule(lr){4-5}
\cmidrule(lr){6-7}
\cmidrule(lr){8-9}
\cmidrule(lr){10-11}

\emph{dropout ratio =} & 0.1\% & 0.5\% & 0.1\% & 0.5\% & 0.1\% & 0.5\% & 0.1\% & 0.5\% & 0.1\% & 0.5\%\\
\midrule
\textbf{GPT2-XL 1.5B}\\
~  \HC\ & \textbf{-34.6\%} & -57.1\% & \textbf{42.9} & 74.0 & -2.4\% & -4.8\% & 2.5 & 5.4 & 0.03 & 0.11 \\
~  \Slim\ & -30.5\% & \textbf{-57.8\%} & 37.7 & \textbf{75.4} & -3.5\% & -6.4\% & 4.1 & 7.5 & 0.02 & 0.17 \\
~  \Zero\ & -29.8\% & -46.1\% & 33.0 & 55.2 & -3.1\% & -4.8\% & 3.5 & 5.5 & 0.03 & 0.09 \\
~  \IG\ & -25.8\% & -40.8\% & 27.0 & 46.0 & \textbf{-2.2\%} & \textbf{-3.4\%} & \textbf{2.3} & \textbf{3.7} & \textbf{0.01} & \textbf{0.05} \\
~  \Act\ & -14.8\% & -29.5\% & 16.9 & 36.4 & -3.0\% & -4.7\% & 3.1 & 5.4 & 0.11 & 0.16 \\
~  \Rand\ & \phantom{1}-0.2\% & \phantom{1}-0.5\% & \phantom{1}0.2 & \phantom{1}0.4 & -0.2\% & -0.5\% & 0.1 & 0.4 & 0.00 & 0.03 \\

\midrule
\textbf{Pythia-deduped 2.8B}\\
~  \HC\ & \textbf{-29.0\%} & \textbf{-53.2\%} & \textbf{55.3} & \textbf{99.8} & -3.7\% & -10.5\% & 7.7 & 22.1 & 0.23 & 0.56 \\
~  \Slim\ & -17.4\% & -45.1\% & 32.9 & 80.8 & -3.3\% & -7.0\% & 6.6 & 13.9 & 0.26 & 0.49 \\
~  \Zero\ & -14.8\% & -25.9\% & 26.4 & 45.2 & -1.1\% & -2.5\% & 2.1 & \phantom{1}5.0 & 0.21 & 0.35 \\
~  \IG\ & -16.7\% & -30.3\% & 29.1 & 52.5 & \textbf{-0.9\%} & \textbf{-2.1\%} & \textbf{1.8} & \phantom{1}\textbf{4.4} & \textbf{0.09} & \textbf{0.18} \\
~  \Act\ & -13.0\% & -25.5\% & 27.5 & 52.2 & -3.1\% & -6.1\% & 6.6 & 12.9 & 0.11 & 0.20 \\
~  \Rand\ & \phantom{1}-0.1\% & \phantom{1}-0.3\% & \phantom{1}0.1 & \phantom{1}0.5 & -0.1\% & -0.3\% & 0.2 & \phantom{1}0.5 & 0.00 & 0.02 \\

\midrule
\textbf{Pythia-deduped 6.9B}\\
~  \HC\ & \textbf{-29.2\%} & \textbf{-57.7\%} & \textbf{58.5} & \textbf{109.9} & -3.8\% & -14.7\% & 8.7 & 32.6 & 0.16 & 0.52 \\
~  \Slim\ & -24.1\% & -48.7\% & 48.8 & 92.1 & -4.2\% & -11.3\% & 9.1 & 23.6 & 0.23 & 0.58 \\
~  \IG\ & -16.9\% & -32.3\% & 31.4 & 57.8 & \textbf{-2.3\%} & \textbf{-4.9\%} & \textbf{5.3} & \textbf{11.5} & 0.27 & \textbf{0.37} \\
~  \Act\ & -11.5\% & -26.8\% & 25.5 & 51.5 & -2.5\% & -8.1\% & 5.5 & 17.2 & \textbf{0.12} & 0.45 \\
~  \Rand\ & \phantom{1}-0.1\% & \phantom{1}-0.2\% & \phantom{1}0.1 & \phantom{1}0.4 & -0.1\% & -0.2\% & 0.1 & \phantom{1}0.3 & 0.00 & 0.02 \\

\bottomrule
\end{tabular}}

\end{center}

\caption{The \BII. \HC\ is the most effective method in erasing the target sequence (\textcolor{blue}{Self}), while \IG\ can best maintain the LLM performance on unrelated sequences (\textcolor{purple}{Neg} and \textcolor{purple}{Rand}) after dropout.}
\label{table:pile}
\end{table*}

%% file: latex/Figs/heatmap.tex
\begin{figure}[t!]
  \centering
  \includegraphics[width=1.\linewidth]{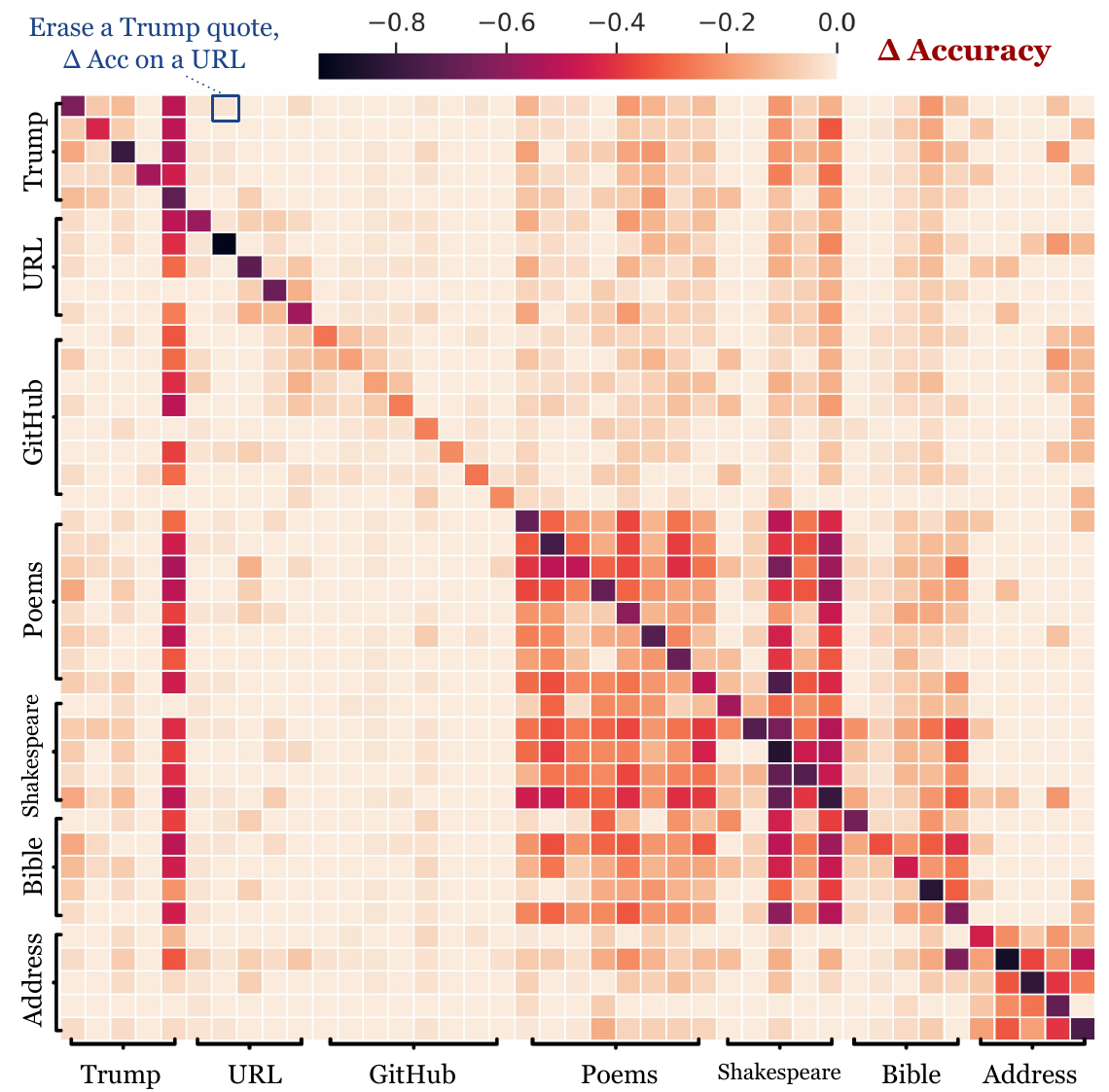}
  \caption{The confusion matrix of \HC\ on a subset of data memorized by GPT2-XL.}
  \label{fig:heatmap}
\end{figure}

%% file: 08_discussion.tex
\section{Related Work and Discussion}
Localization identifies function-specific components, including neurons \cite{radford2017learning,gurnee2023finding}, layers \cite{gupta2023editing}, or subnetworks \cite{csordas2020neural,cao-etal-2021-low,foroutan-etal-2022-discovering}.
For example, \citet{dai-etal-2022-knowledge} find knowledge neurons for each relational fact.
\citet{meng2022locating} locate relational facts to middle FFNs, specifically when LLMs process the last token of the subject.
\citet{bayazit2023discovering} discover sparse knowledge subnetworks in GPT2 with a differentiable weight masking method.
However, there is no standard approach to evaluate the effectiveness of localization methods.
We are the first to systematically and directly compare different methods on LLMs of different sizes, including knowledge neurons (\IG) and differentiable masking methods \Slim\ and \HC.

We take the view that LLM memorization of a sequence is different from learning a type of knowledge.
Memorization is reproducing a long sequence (near) verbatim. 
In contrast, knowledge, often represented as a <subject, relation, object> triplet, occurs in variable contexts, where paraphrases are treated as equivalent expressions of the same knowledge.
Localization of verbatim memorization helps unlearn private or copyrighted data, for example, \citet{wu-etal-2023-depn} apply \IG\ to localize and then erase private data from a BERT fine-tuned on Enron Email dataset \cite{klimt2004introducing}.
Our \BII\ differs from \citet{wu-etal-2023-depn} in two main ways: (1) we delete sequences that LLMs have naturally memorized during pretraining, (2) we locate neurons for each sequence independently, rather than finding a shared set of neurons, as our collected datasets cover diverse sequences.
Localization can also prevent overfitting: \citet{maini2023can} drop out pre-allocated neurons tied to memorizing mislabeled examples.
In contrast with these works, we focus on benchmarking localization ability, since successful localization is the basis of its downstream applications.

%% file: latex/Figs/ig-single-layer.tex
\begin{figure}[t!]
  \centering
  \includegraphics[width=0.9\linewidth]{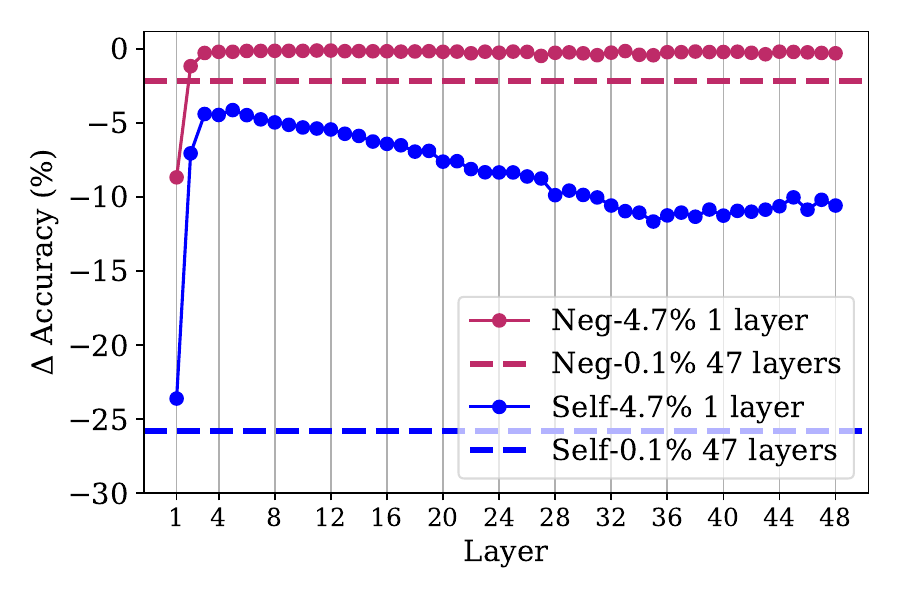}
  \caption{Dropout in one layer vs. multiple layers.}
  \label{fig:ig-single}
\end{figure}

%% file: 09_conclusion.tex
\section{Conclusion}
We propose two benchmarking approaches to define the success of LLM localization, focusing on locating a small set of neurons in an LLM that are responsible for memorizing a sequence.
The \BI\ enables a direct evaluation of localization methods, while the \BII\ evaluates methods on naturally memorized sequences, using dropout to measure localization success.
The two benchmarks complement each other and show consistent rankings of methods.
We find promising localization ability of all five methods we evaluate, especially for \HC.
Meanwhile, all methods confuse memorized sequences in the same or related categories.
This finding suggests a need for better localization methods and poses the open question of whether LLMs use a shared set of neurons to memorize related sequences such that perfect localization is not possible.

%% file: 10_limitations.tex
\section{Limitations}
We follow prior work (\cref{sec: ffn-background}) and assume that FFNs are the most important components in LLMs for memorizing data; thus, we only study localization in FFNs, not considering other model components such as attention layers.
Similarly, we focus on neurons instead of individual weights in FFNs, so as to make fair comparisons with existing methods, \IG\ and \Act.

In the \BI, we assume that all the fine-tuned weights are responsible for memorizing the newly injected sequence.
However, there is no guarantee that all of them contribute to memorization.
We roughly address this issue by lowering the injection ratio, which makes it less likely for the model to memorize the injected sequence without using all of the chosen weights; indeed, we observe that when the ratio is $10 \times$ smaller, all localization methods achieve higher recalls in Table \ref{table:ecbd}, even though the random baseline performs $10\times$ worse.

We acknowledge the limitations of evaluating localization in our \BII.
First, we use dropout (namely, zero ablation) to observe the importance of the located neurons, which is only one possible way to approach localization; other approaches such as mean ablation \cite{wang2023interpretability} and path patching \cite{goldowsky2023localizing,hanna2023how} are not covered in this paper.
Besides, given a target sequence, we treat all the other memorized sequences as its negative examples without considering semantic overlap or data sources, as our data deduplication only ensures there is little lexical overlap between sequences (\cref{sec:B2}).
However, we find all localization methods show confusion between several quotes, which may share semantic similarities or co-occur in some pretrained documents.
It is debatable whether related examples should be considered negative, and it depends on what the goal of localization is.
We invite future work to propose new ways to define the success of localization for the \BII.

%% file: acknowledgements.tex
\section*{Acknowledgements}
We thank Johnny Wei, Eric Wallace, and Ameya Godbole for their help in finding memorized data.
We thank Cheng-Han Chiang for helpful discussions on model editing.
We thank Wang Zhu, Ting-Rui Chiang, Joshua Robinson, Lee Kezar, Deqing Fu, Anthony Liang, Ta-Chung Chi, Yi-Lin Tuan, Yau-shian Wang, and the anonymous reviewers for their valuable feedback.
We thank USC NLP cluster admin for their great work on keeping the servers functional.
This work was funded in part by gifts from Open Philanthropy and Cisco Research.

%% file: 11_appendix.tex
\section{Appendix}
\label{sec:appendix}

\subsection{The Loss for Information Injection}
\label{sec:app_inj}
In the \BI, we use regular language modeling loss to train the LLM $\LLM$ on a new sequence $x = [x_1, \dotsc, x_\T]$ of $\T$ tokens.
Formally,
\begin{equation}
    \frac{1}{\T-1}\sum_{t=2}^{\T} - \log \LMP (x_t | x_{ < t}) \label{eq: lm_loss}
\end{equation}
Here, the index $t$ starts from $2$, because all the LLMs we use (GPT2 and Pythia models) do not add <bos> tokens to data when doing language modeling in their pretraining.
Therefore, there is no loss on the first token $x_1$ and the total loss is averaged across $\T - 1$ token.

\subsection{Details of \IG}
\label{sec:app_ig}
Recall that a sequence $x=(p, s)$ consists of a prefix $p$ and a suffix $s = [s_1, \dotsc, s_T]$.
Denote $\IGP$ as the LLM output probability of token $s_t$ if we replace the original hidden state at the $l$-th layer, $\OH \in \mathbb{R}^{d_2}$, with a new hidden state $\NH \in \mathbb{R}^{d_2}$:
\begin{equation}
    \IGP = \LMP(s_t|p, s_{<t}, \NH)
\end{equation}
To calculate the integrated gradients along the $i$-th neuron dimension, we gradually change $\NH$ from a zero vector\footnote{We follow \citet{dai-etal-2022-knowledge} to set the \emph{baseline} in integrated gradients to a zero vector that has the same shape as $\OH$.} to the original hidden state $\OH$, and cumulating the gradients of $\operatorname{P}(\cdot)$ along the $i$-th dimension.
Finally, we get the attribution score $\Attr$ by averaging the integrated gradients across the suffix length $T$:
\begin{align}
\textbf{IG}_i(z) &:= \; z_i \, \int_{\alpha=0}^{1} \frac{\partial \p(\alpha z)}{\partial z_i} d \alpha, \label{eq: integral} \\
\Attr &= \frac{1}{T}\sum_{t=1}^{T}\textbf{IG}_i(h_t^l)
\end{align}
where $\textbf{IG}_i(h_t^l)$ is the integrated gradients along the $i$-th neuron dimension in the $l$-th layer at the $t$-th timestep, when the input is $[p, s_{<t}]$.
\citet{sundararajan2017axiomatic} compute Riemann sum to approximate Eq. \ref{eq: integral}, which uses a fixed number of intervals to approximate the integrals.
We closely follow the implementation of \url{https://github.com/EleutherAI/knowledge-neurons}.

\subsection{Details of \Slim}
We initialize every mask value $\Mask$ as $1$, which is equivalent to running the pretrained LLM without masking.
When training the mask, we clip every $\Mask$ to $[0, 1]$.
Note that for both \Slim\ and \HC, because we are learning a mask on each neuron, we do not apply any random dropout during training.

\subsection{Details of \HC}
\label{sec:app_HC}
\citet{louizos2017learning} obtain the hard concrete r.v. $\ClipMask_i$ by first stretching the binary concrete r.v. $\HCMask$ (Eq. \ref{eq:bin_rv}) from the interval $(0, 1)$ to $(\gamma , \zeta)$, where $\gamma = -0.1, \zeta = 1.1$, and then clip the value to the $[0, 1]$ interval:
\begin{equation}
    \ClipMask_i = \min \left( 1, \max \left(0,\, \HCMask\ \cdot (\zeta - \gamma) + \gamma \right) \right) \nonumber 
\end{equation}
They then use $L_0$ regularization to encourage sparsity on the weights after applying the mask $\ClipMask$.
After reparametrization, they have the regularization $\Reg$:
\begin{equation}
\Reg\ = \sum_{i=1}^{d_2} \sigma \left(\log m^l_i - C \right), 
\end{equation}
where $C = \beta \log \frac{-\gamma}{\zeta}$ is a constant.

\subsection{Expanding the dataset of \BI}
\label{sec:app_more_data}
We double the data size of the \BI\ by including the newly released ECBD 2022, 2023 splits, having 328 distinct definition sentences from ECBD 2021-2023. 
We experiment with this expanded dataset on GPT2, injection ratio=$0.1\%$, using the same hyperparameters as Table \ref{table:ecbd}.
Table \ref{tab:app_expand} shows the results on ECBD 2021-2023 are very close to the ones on ECBD-2021 only (Table \ref{table:ecbd}), suggesting that our conclusions hold when we increase the dataset size.
\input{latex/Tables/app_expand}
 
\subsection{Do random seeds affect the results of the \BI?}
\label{sec:app_more_seeds}
In \BI, we sample different sets of weights for different examples (see Algorithm \ref{algo:inj}); thus, the results reported in Table \ref{table:ecbd} are averaged over many different choices of weights.
To further show that random seeds do not affect our results, we run an additional experiment on GPT2, with the injection ratio=$0.1\%$.
Specifically, for each example, we choose a different random seed and thus choose a different set of weights to inject the example.
Comparing the new results in Table \ref{tab:app_seeds} with the original ones in Table \ref{table:ecbd}, we find that the recall scores barely change for all localization methods. 
Also, for each method, we run paired two-tailed t-tests comparing the recalls between the original and new seeds and observe that all pairs have p-values $> 0.05$, suggesting that differences between random seeds are not significant.
\input{latex/Tables/app_seeds}

\input{latex/Tables/time}

\subsection{Computation costs of different methods}
\label{sec:app_cost}
Among all five localization methods, \Act\ is the most computationally efficient, because Eq. \ref{eq:act} only requires one forward pass.
Both the pruning-based methods \Slim\ and \HC\ perform fast, as only the masks are trainable.
Calculating integrated gradients (\IG) is time-consuming, while \Zero\ is the worst, because it leaves out every neuron one by one.
We compare the computational cost of different methods on one sequence memorized by Pythia-deduped-6.9B, where each sequence in the collected set $\mathcal{X}$ consists of a 32-token prefix and a 48-token suffix.
We follow the common implementation that sets the number of intervals to $20$ for Riemann sum in \IG.
Table \ref{tab:time} shows the elapsed time calculated on an RTX A6000 48G GPU.
When running \IG\ and \Zero\, we patch and batch the activations to reach $99\%$ GPU utilities.
Still, applying \Zero\ to do localization on one sequence costs 8.5 hours, and $\mathcal{X}$ contains 500 sequences in total.
Due to the extremely heavy computation cost, we do not have the results of \Zero\ on Pythia-6.9B in the \BII.

\subsection{Details of Data Collection}
\label{sec:app_data}
We show some collected examples in Tables \ref{tab:app_exampels}\&\ref{tab:app_exampels2}.
Table \ref{tab:before} reports how well the pretrained LLMs memorize sequences in the collected datasets.
\input{latex/Tables/before}

\paragraph{The pretrained sequences of Pythia models.}
EleutherAI releases the exact batches used by Pythia models during pretraining, where each sequence in a batch consists of $2049$ tokens \footnote{\url{https://github.com/EleutherAI/pythia\#exploring-the-dataset}}.
We first randomly downsample the pretraining batches to a subset $\mathcal{Z}$ of $102400$ sequences.
Then, we use our criteria in \cref{sec:B2} to determine whether Pythia memorizes a sequence in the subset.
After filtering, there remain $500 \sim 1000$ sequences in the subsets for both Pythia-deduped-2.8B and Pythia-deduped-6.9B; we simply sample $505$ of them respectively as our collected datasets.

We also randomly sample a subset of $2048$ sequences ($\mathcal{D}$), each consisting of $128$ tokens, to measure the perplexity of all LLMs we evaluate.
We ensure that $\mathcal{Z} \cap \mathcal{D} = \varnothing$, so there is no overlap between the collected memorized sequences and sequences for perplexity.

\paragraph{Filtering with greedy decoding.}
Given the prefix $p$ as the prompt to the LLM, we generate the suffix $\bar{s} = [\bar{s_1}, \dotsc, \bar{s_{48}}]$ with greedy decoding, where
\begin{equation}
\bar{s_t} = \argmax_{\Tok \in \text{Voc}} \, \LMP (\Tok | p, \bar{s}_{<t}). \label{eq:s_bar}
\end{equation}
We then calculate the Levenshtein distance \cite{Levenshtein1965BinaryCC} between the true suffix $s$ and the generated one $\bar{s}$, filtering out sequences with a distance greater than $20$ characters.
Note $\bar{s}$ is different from $\hat{s}$ in Eq \ref{eq:argmax}, which is generated by teacher-forcing and is used to calculate memorization scores.

\paragraph{Deduplication.}
Although we use the deduplicated version of the dataset and models, the Pile-dedupe and Pythia-deduped models, we still find lots of near-duplicated sequences.
Thus, we further deduplicate the collected memorized sequences.
In particular, we follow \citet{lee-etal-2022-deduplicating} to represent each sequence with a set of 5-grams when calculating the Jaccard index.
Among a set of duplicates, we select the one that is best memorized, i.e., having the lowest Levenshtein distance on the generated suffix $\bar{s_t}$ (Eq. \ref{eq:s_bar}), and discard the others.
\input{latex/Figs/single}
\paragraph{Manually searched data.}
With our searching criteria in \cref{sec:B2}, we can only identify less than 10 memorized sequences from subsets of the Pile-dedupe, Common Crawl, and Wikipedia, probably because OpenAI carefully preprocesses the data before training GPT2-XL.
Thus, we actively search for potentially memorized data, such as famous poems and common lists of sorted items.
We collect 105 sequences memorized by GPT2-XL and manually categorize them (see Tables \ref{tab:exampels} \& \ref{tab:app_exampels}), including 31 examples from \citet{carlini2021extracting}.
We set the prefix and suffix of a sequence by trial and error to ensure high memorization Accuracy.
Unlike automatic searches that tend to find templated texts or uninteresting data with repetitive tokens \cite{zhang2021counterfactual}, our manual search ensures better data quality and enables us to analyze memorization within and across categories.

In particular, Figures \ref{fig:app_heatmap_all} \& \ref{fig:app_heatmap_hc} show that different localization methods constantly confuse sequences of related categories.
For example, they are unable to disentangle neurons of different quotes and identify a small set of neurons responsible for both the order of Zodiac Signs and the order of Planets.

\paragraph{Responsible checklist.}
Note the \texttt{Contact Info} category of our manually collected dataset only contains public data, such as mailing addresses of corporate headquarters and famous buildings; thus, it does not have any potential risk of revealing private information.
Similarly, our memorized datasets for Pythia models are a subset of the Pile, a public corpus under the MIT License.

\subsection{Hyperparameters}
\label{sec:app_hyper}
In the \BI, the ECBD-2021 set contains $156$ definition sequences.
For the \BII, we collect a set of 505, 505, and 105 sequences memorized by Pythia-deduped-6.9B, Pythia-deduped-2.8B, and GPT2-XL, respectively.
For each set, we sample 5 sequences as the dev set, using the dev set performance to determine the hyperparameters for each LLM.
The hyperparameters include the integrated gradient steps, i.e., the number of intervals in Riemann sum for integral approximation in \IG; the temperature $\beta$ and the initialization value of parameters $m$ in \HC; the learning rate, the number of training epochs, and $\lambda$, which balances the memorization loss and the sparsity loss, in \Slim\ and \HC.
We observe that both \Slim\ and \HC\ are sensitive to the choice of hyperparameters.
On the other hand, we find the performance of \IG\ does not improve when using more integrated gradient steps, where we experiment with different steps ranging from 20 to 300.
Thus, we set the step to 20 for all examples to reduce the heavy computation costs.

\subsection{More experiments comparing dropout in one layer with multiple layers.}
\label{sec:single}
In \cref{sec: main_single}, neurons are localized by \IG.
In this section, we conduct the same experiment using \Zero\ and \Act\ methods.
Figure \ref{fig:single} shows that dropping out $N$ neurons in multiple layers (Self-$0.1\%$ 47 layers) even outperforms dropping out $5 \times N$ neurons in a single layer (Self-$23.5\%$ 1 layer), except for the bottom layers where the memorization of both target and negative examples are greatly hurt. 
Hence, we believe the memory of a piece of information is distributed across layers; meanwhile, only a few weights in each layer are mainly responsible for the memorization (\cref{sec:b2_res}).

We do not have the single-layer results of \Slim\ and \HC, because both methods train the masks of all neurons jointly, which requires us to retrain the masks only on a single layer to obtain their attribution scores.
In comparison, the other three methods consider each neuron individually, allowing us to use the same attribution scores to select neurons in a single layer and make direct comparisons with the results in Table \ref{table:pile} (the dashed lines in Figure \ref{fig:single}).

\subsection{Predicting top neurons across layers}
\label{sec:app_global}
In the \BI, we randomly sample weights across layers to inject the data, instead of sampling a fixed percentage of weights per layer (see Algorithm \ref{algo:inj}).
Hence, it may seem more natural to predict top-$k\%$ of neurons across layers; we experiment with this alternative in Table \ref{table:ecbd_global}.

Comparing the results of Table \ref{table:ecbd} and Table \ref{table:ecbd_global}, we find that predicting top neurons per layer outperforms predicting top neurons across layers.
This is because all localization methods assign larger attribution scores to neurons in the bottom layers, barely predicting neurons in the upper layers if we rank neurons globally.
On the other hand, Table \ref{table:ecbd} and Table \ref{table:ecbd_global} show consistent results.
Our findings and the ranking of different methods are coherent whether we rank neurons per layer or globally.

\subsection{Implementation Details}
Table \ref{tab:archi} summarizes the architectures of LLMs we use.
We run most experiments on RTX3090 24G GPUs; experiments involving Pythia-6.9B are run on RTXA6000 48G GPUs.
We use \texttt{transformers 4.31.0} and \texttt{pytorch 1.13}.
\input{latex/Tables/app_archi}

\input{latex/Tables/ECBD-global}

\input{latex/Tables/app_pvalues_inj}
\input{latex/Tables/app_pvalues_del}

\input{latex/Tables/app_examples}
\input{latex/Tables/app_examples2}

\input{latex/Figs/appendix_heatmap_all}
\input{latex/Figs/appendix_heatmap_HC}

%% file: latex/Tables/app_expand.tex
\begin{table}[!h]
\small
\begin{center}
\centering
{
\begin{tabular}{lccc}
\toprule
& \multicolumn{3}{c}{\textbf{GPT2 124M}} \\
\midrule
\emph{ratio = 0.1\%} & \textbf{R@0.1\%} & \textbf{R@0.2\%} & \textbf{R@0.5\%} \\
\cmidrule(lr){2-4}

~  HC & $58.3$ & $81.6$ & $\textbf{94.6}$ \\
~  Slim & $\textbf{59.2}$ & $\textbf{84.3}$ & $94.5$ \\
~  IG & $53.3$ & $73.0$ & $84.1$ \\
~  Activation & $11.1$ & $26.5$ & $52.7$ \\
\bottomrule

\end{tabular}}

\end{center}

\caption{The \BI\ results of GPT2 on the expanded dataset, ECBD 2021-2023.}
\label{tab:app_expand}
\end{table}

%% file: latex/Tables/app_seeds.tex
\begin{table}[!h]
\small
\begin{center}
\centering
{
\begin{tabular}{lccc}
\toprule
& \multicolumn{3}{c}{\textbf{GPT2 124M}} \\
\midrule
\emph{ratio = 0.1\%} & \textbf{R@0.1\%} & \textbf{R@0.2\%} & \textbf{R@0.5\%} \\
\cmidrule(lr){2-4}

~  HC & $57.9$ & $81.0$ & $94.6$ \\
~  Slim & $\textbf{59.6}$ & $\textbf{84.4}$ & $\textbf{94.8}$ \\
~  IG & $54.0$ & $73.7$ & $83.9$ \\
~  Activation & $11.4$ & $26.4$ & $52.7$ \\
\bottomrule

\end{tabular}}

\end{center}

\caption{The \BI\ results with a new set of random seeds. The Recall@$k\%$ scores are very similar to the original ones in Table \ref{table:ecbd}, showing the \BI\ is not sensitive to the choice of random seed.}
\label{tab:app_seeds}
\end{table}

%% file: latex/Tables/time.tex
\begin{table}[!b]
\begin{center}
\centering
\resizebox{0.75\columnwidth}{!}{
\begin{tabular}{|l|l|}
\hline
& \phantom{3} \textbf{Time} \\
\hline
\Act\ & $\sim 0.3$ sec  \\
\Slim\ & $\sim 12$ sec \\
\HC\ &  $\sim 1$ min \\
\IG\ &  $\sim 43$ min \\
\Zero\ &  $\sim 8.5$ hr \\
\hline
\end{tabular}}
\end{center}
\caption{The elapsed time of different methods to do localization (i.e., assign attribution scores to every neuron) on one sequence memorized by Pythia-6.9B. We time all methods on a single RTX A6000 GPU.}
\label{tab:time}
\end{table}

%% file: latex/Tables/before.tex
\begin{table}[!h]
\small
\begin{tabular}{lcccc}
\toprule
& \textbf{Acc} & \textbf{Dist} & \textbf{PPL} & Len \\
\bottomrule
\textbf{GPT2-XL} & 99.3\% & 0.48 & 10.18 & 150 \\
\textbf{Pythia-deduped-2.8B} & 98.8\% & 1.07 & \phantom{1}5.58 & 160 \\
\textbf{Pythia-deduped-6.9B} & 99.7\% & 0.20 & \phantom{1}5.24  & 167\\
\bottomrule
\end{tabular}
\caption{Quantifying memorization of the collected datasets. The high Accuracy (Acc) and low Levenshtein distance (Dist) show our collected sequences ($\mathcal{X}$) are indeed well memorized by LLMs. The last column (Len) reports the average suffix length of each dataset at the character level. We also measure the perplexity (PPL) on sequences sampled from the Pile-dedupe ($\mathcal{D}$).}
\label{tab:before}
\end{table}

%% file: latex/Figs/single.tex
\begin{figure*}[th]
  \centering
  \includegraphics[width=1.\linewidth]{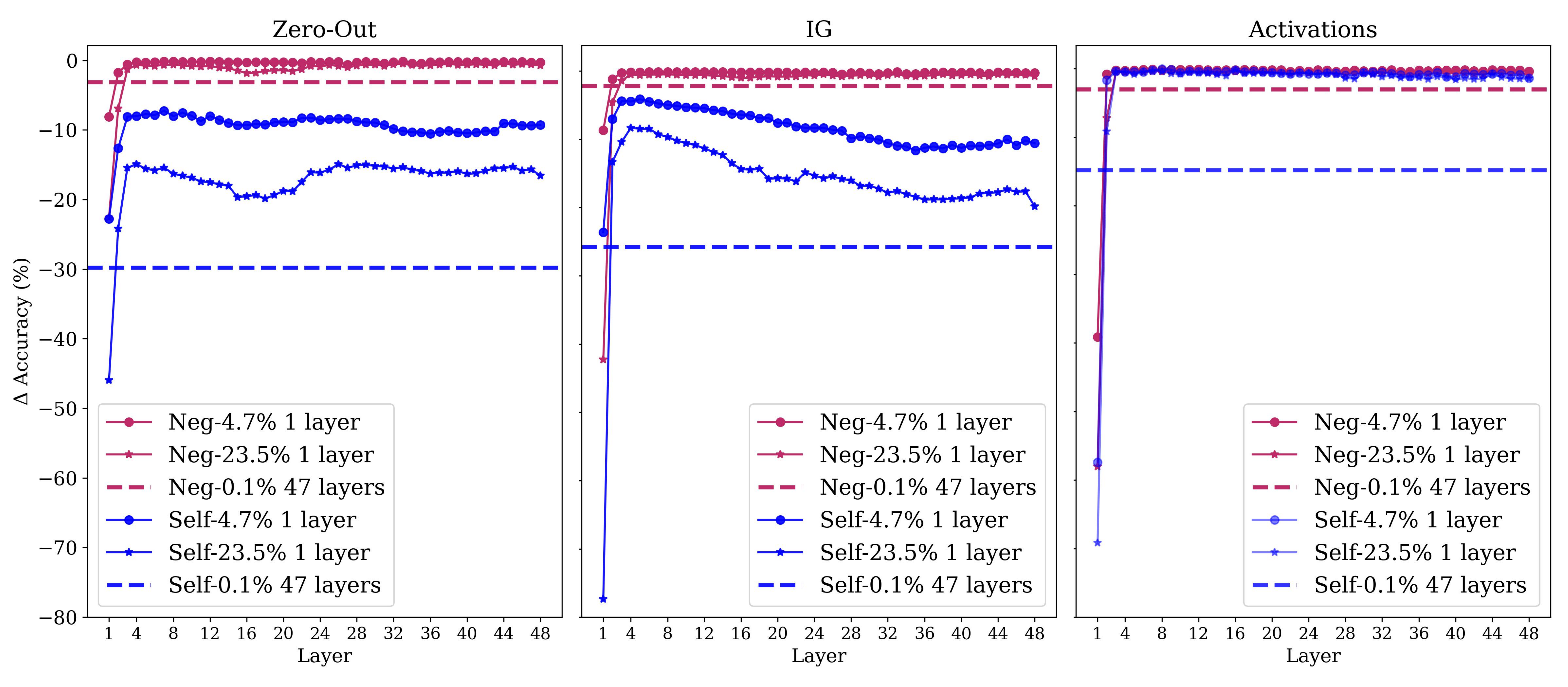}
  \caption{The \BII\ results of \Zero, \IG, and \Act\ methods when dropping out the same number of neurons in a single layer, where the blue lines show \textcolor{blue}{$\Delta$ Self-Acc} and the red lines show \textcolor{purple}{$\Delta$ Neg-Acc}. Under the same ``neuron budget’’, dropping out neurons in multiple layers (blue dashed lines) substantially outperforms dropout in a single layer, implying that memorized information is stored in a distributed fashion over multiple layers. Besides, dropping out neurons in the bottom layer greatly hurts the memorization of negative examples (red lines), suggesting that the bottom layer encodes general information.}
  \label{fig:single}
\end{figure*}

%% file: latex/Tables/app_archi.tex
\begin{table}[!h]
\begin{center}
\centering
\begin{tabular}{lccc}
\toprule
& \textbf{\# Layers} & \textbf{\# Neurons}  \\
\bottomrule
GPT2 124M & 12& \phantom{1}3072 \\
GPT2-XL 1.5B & 48 & \phantom{1}6400 \\
Pythia-deduped-2.8B & 32 & 10240 \\
Pythia-deduped-6.9B & 32 & 16384 \\
\bottomrule
\end{tabular}
\end{center}
\caption{The number of layers and the number of FFN neurons in each layer of different LLMs.}
\label{tab:archi}
\end{table}

%% file: latex/Tables/ECBD-global.tex
\begin{table*}[!t]
\begin{center}
\centering
\resizebox{2.\columnwidth}{!}
{
\begin{tabular}{lcccccccccccc}
\toprule
      & \multicolumn{3}{c}{\textbf{GPT2 124M}} 
      & \multicolumn{3}{c}{\textbf{GPT2-XL 1.5B}} 
      & \multicolumn{3}{c}{\textbf{Pythia-deduped 2.8B}} 
      & \multicolumn{3}{c}{\textbf{Pythia-deduped 6.9B}}\\

\cmidrule(lr){2-4}
\cmidrule(lr){5-7}
\cmidrule(lr){8-10}
\cmidrule(lr){11-13}

& R@1\% & R@2\% & R@5\% & R@1\% & R@2\% & R@5\% & R@1\% & R@2\% & R@5\% & R@1\% & R@2\% & R@5\%\\
\midrule
\emph{ratio = 1\%}\\
~  \HC\ & $\textbf{46.6}$ & $\textbf{66.8}$ & $\textbf{88.0}$ & $\textbf{21.8}$ & $\textbf{25.1}$ & $\textbf{32.8}$ & $33.3$ & $\textbf{48.4}$ & $\textbf{70.7}$ & $31.5$ & $47.5$ & $\textbf{69.4}$ \\
~  \Slim\ & $43.1$ & $64.6$ & $79.9$ & \phantom{1}$5.2$ & $11.5$ & $27.0$ & $\textbf{33.6}$ & $47.3$ & $59.8$ & $\textbf{35.0}$ & $\textbf{49.6}$ & $63.4$ \\
~  \Zero\ & $24.0$ & $36.8$ & $52.7$ & \phantom{1}$4.2$ & \phantom{1}$7.3$ & $13.5$ & $10.1$ & $14.3$ & $20.5$ & - & - & - \\
~  \IG\ & $10.3$ & $18.1$ & $36.3$ & \phantom{1}$1.4$ & \phantom{1}$4.8$ & $12.2$ & \phantom{1}$6.1$ & $10.8$ & $21.1$ & \phantom{1}$8.9$ & $13.9$ & $24.1$ \\
~  \Act\ & \phantom{1}$2.5$ & \phantom{1}$4.4$ & \phantom{1}$9.8$ & \phantom{1}$1.5$ & \phantom{1}$2.8$ & \phantom{1}$6.8$ & \phantom{1}$3.2$ & \phantom{1}$5.1$ & $21.6$ & \phantom{1}$4.1$ & \phantom{1}$6.3$ & $17.4$ \\
~  \Rand\ & \phantom{1}$1.0$ & \phantom{1}$2.0$ & \phantom{1}$5.0$ & \phantom{1}$1.0$ & \phantom{1}$2.0$ & \phantom{1}$5.0$ & \phantom{1}$1.0$ & \phantom{1}$2.0$ & \phantom{1}$5.0$ & \phantom{1}$1.0$ & \phantom{1}$2.0$ & \phantom{1}$5.0$ \\

\midrule

\emph{ratio = 0.1\%} & @0.1\% & @0.2\% & @0.5\% & @0.1\% & @0.2\% & @0.5\% & @0.1\% & @0.2\% & @0.5\% & @0.1\% & @0.2\% & @0.5\%\\
\cmidrule(lr){2-4}
\cmidrule(lr){5-7}
\cmidrule(lr){8-10}
\cmidrule(lr){11-13}
\\
~  \HC\ & $51.2$ & $77.4$ & $\textbf{96.4}$ & $\textbf{49.8}$ & $\textbf{57.5}$ & $\textbf{63.6}$ & $\textbf{45.6}$ & $\textbf{65.5}$ & $\textbf{85.9}$ & $28.7$ & $40.7$ & $55.8$ \\
~  \Slim\ & $\textbf{62.7}$ & $\textbf{87.0}$ & $95.4$ & $18.1$ & $35.1$ & $54.0$ & $45.0$ & $62.6$ & $73.6$ & $\textbf{39.1}$ & $\textbf{52.1}$ & $\textbf{64.3}$ \\
~  \Zero\ & $57.4$ & $81.7$ & $91.9$ & $14.7$ & $20.9$ & $31.1$ & $16.4$ & $20.6$ & $25.8$ & - & - & - \\
~  \IG\ & $36.0$ & $55.0$ & $75.5$ & \phantom{1}$2.5$ & \phantom{1}$3.5$ & \phantom{1}$6.0$ & $12.6$ & $16.4$ & $21.9$ & $19.7$ & $23.6$ & $28.9$ \\
~  \Act\ & \phantom{1}$9.0$ & $12.9$ & $23.4$ & \phantom{1}$3.5$ & \phantom{1}$4.6$ & \phantom{1}$6.7$ & \phantom{1}$8.0$ & $16.8$ & $40.5$ & $21.2$ & $31.4$ & $50.2$ \\
~  \Rand\ & \phantom{1}$0.1$ & \phantom{1}$0.2$ & \phantom{1}$0.5$ & \phantom{1}$0.1$ & \phantom{1}$0.2$ & \phantom{1}$0.5$ & \phantom{1}$0.1$ & \phantom{1}$0.2$ & \phantom{1}$0.5$ & \phantom{1}$0.1$ & \phantom{1}$0.2$ & \phantom{1}$0.5$ \\

\bottomrule
\end{tabular}}

\end{center}

\caption{The \BI. The average Reacall@$k\%$ of different methods when predicting top-$k\%$ of neurons \emph{across} layers. The results are consistent with Table \ref{table:ecbd}, where methods predict a fixed $k\%$ of neurons in each layer.}
\label{table:ecbd_global}
\end{table*}

%% file: latex/Tables/app_pvalues_inj.tex
\begin{table*}[!t]
\begin{center}
\centering
\resizebox{2.\columnwidth}{!}
{
\begin{tabular}{lcccccccccccc}
\toprule
      & \multicolumn{3}{c}{\textbf{GPT2 124M}} 
      & \multicolumn{3}{c}{\textbf{GPT2-XL 1.5B}} 
      & \multicolumn{3}{c}{\textbf{Pythia-deduped 2.8B}} 
      & \multicolumn{3}{c}{\textbf{Pythia-deduped 6.9B}}\\

\cmidrule(lr){2-4}
\cmidrule(lr){5-7}
\cmidrule(lr){8-10}
\cmidrule(lr){11-13}

& R@1\% & R@2\% & R@5\% & R@1\% & R@2\% & R@5\% & R@1\% & R@2\% & R@5\% & R@1\% & R@2\% & R@5\%\\
\midrule
\emph{injection ratio = 1\%}\\
~   \HC & $9E-85$ & $7E-86$ & $2E-77$ & $2E-100$ & $8E-108$ & $8E-111$ & $3E-114$ & $3E-121$ & $6E-138$ & $5E-111$ & $6E-129$ & $3E-155$ \\
~ \Slim & $1E-72$ & $8E-73$ & $4E-62$ & $1E-66$ & $5E-77$ & $2E-84$ & $3E-111$ & $5E-119$ & $4E-119$ & $1E-121$ & $3E-134$ & $2E-132$ \\

\midrule

 & @0.1\% & @0.2\% & @0.5\% & @0.1\% & @0.2\% & @0.5\% & @0.1\% & @0.2\% & @0.5\% & @0.1\% & @0.2\% & @0.5\%\\
\cmidrule(lr){2-4}
\cmidrule(lr){5-7}
\cmidrule(lr){8-10}
\cmidrule(lr){11-13}
\emph{injection ratio = 0.1\%} \\
~  \HC & \textcolor{red}{$6E-03$} & $9E-06$ & $1E-18$ & $2E-96$ & $3E-87$ & $6E-73$ & $7E-91$ & $7E-110$ & $1E-127$ & $2E-42$ & $7E-77$ & $3E-83$ \\
~   \Slim & $3E-10$ & $4E-17$ & $2E-20$ & $1E-62$ & $2E-73$ & $4E-71$ & $6E-96$ & $7E-101$ & $4E-99$ & $1E-52$ & $1E-54$ & $1E-52$ \\

\bottomrule
\end{tabular}}

\end{center}

\caption{The p-values of the \BI. $H_0$: \IG\ and pruning-based methods, \HC\ or \Slim, have identical expected Recall@$k\%$ scores on ECBD 2021 examples.
As we have 24 settings in total, we run 24 one-tailed paired t-tests with Bonferroni correction, setting the significance level $\alpha=\frac{0.05}{24}$. We color the results that have p-values $>\alpha$.}
\label{table:app_pvalues_inj}
\end{table*}

%% file: latex/Tables/app_pvalues_del.tex
\begin{table*}[!t]
\begin{center}
\centering
\small
{
\begin{tabular}{lcccccccccccc}
\toprule
      & \multicolumn{2}{c}{\textbf{GPT2-XL 1.5B}} 
      & \multicolumn{2}{c}{\textbf{Pythia-deduped 2.8B}} 
      & \multicolumn{2}{c}{\textbf{Pythia-deduped 6.9B}}\\

\cmidrule(lr){2-3}
\cmidrule(lr){4-5}
\cmidrule(lr){6-7}

\emph{dropout ratio =} & 0.1\% & 0.5\% & 0.1\% & 0.5\% & 0.1\% & 0.5\%\\
\midrule

~  \HC\ & $1.6E-10$ & $3.8E-17$ & $5.3E-61$ & $4.1E-78$ & $2.2E-61$ & $3.2E-90$ \\
~  \Slim\ & $1.6E-04$ & $1.8E-15$ & \textcolor{red}{$1.2E-01$} & $8.9E-55$ & $2.4E-24$ & $2.0E-60$ \\ 

\bottomrule
\end{tabular}}

\end{center}

\caption{The p-values of the \BII, where we focus on the memorization accuracy of the target examples. $H_0$: \IG\ and pruning-based methods, \HC\ or \Slim, have identical expected \textcolor{blue}{$\Delta$ Self-Acc} scores on the memorized sequences. As we have 6 settings in total, we run 6 one-tailed paired t-tests with Bonferroni correction, setting the significance level $\alpha=\frac{0.05}{6}$. We color the results that have p-values $>\alpha$.}
\label{table:app_pvalues_del}
\end{table*}

%% file: latex/Tables/app_examples.tex
\begin{table*}[!t]
\begin{center}
\begin{tabular}{|p{2.5cm}|p{0.8cm}|p{11cm}|}
\hline
Email & 100\% & \textcolor{brown}{Write to Jon Hilsenrath at} jon.hilsenrath@wsj.com \\
\hline
Zodiac Signs & 100\% & \textcolor{brown}{Aries Taurus} Gemini Cancer Leo Virgo Libra Scorpio Sagittarius Capricorn Aquarius Pisces \\
\hline
Patreon & 100\% & \textcolor{brown}{Thank you to our Patreon Supporters: Saintsofwar, Anon,} Lord\_Of\_Fapping, Dryzak, Chabalbac, ioNz, LaX, VNT \\
\hline
Declaration of Independence & 100\% & \textcolor{brown}{We hold these truths to be self-evident, that all} men are created equal, that they are endowed by their Creator with certain unalienable Rights, that among these are Life, Liberty and the pursuit of Happiness. \\
\hline
Trump & 100\% & \textcolor{brown}{Sorry losers and haters, but my I.Q. is one} of the highest -and you all know it! Please don't feel so stupid or insecure, it's not your fault. \\
\hline
Newton & 100\% & \textcolor{brown}{I do not know what I may appear to the world, but to myself I seem to have been only like a boy} playing on the sea-shore, and diverting myself in now and then finding a smoother pebble or a prettier shell than ordinary, whilst the great ocean of truth lay all undiscovered before me. \\
\hline
Dr. MLK & 100\% & \textcolor{brown}{And when this happens, and when we allow freedom ring, when we let it ring from every village and every hamlet,} from every state and every city, we will be able to speed up that day when all of God's children, black men and white men, Jews and Gentiles, Protestants and Catholics, will be able to join hands and sing in the words of the old Negro spiritual, "Free at last! Free at last! Thank God Almighty, we are free at last" \\
\hline
Genesis & 100\% & \textcolor{brown}{In the beginning God created the heaven and the earth. And} the earth was without form, and void; and darkness was upon the face of the deep. And the Spirit of God moved upon the face of the waters. And God said, Let there be light: and there was light. \\
\hline
The Road Not Taken & 100\% & \textcolor{brown}{Two roads diverged in a yellow wood,}\textbackslash n\textbackslash nAnd sorry I could not travel both\textbackslash n\textbackslash nAnd be one traveler, long I stood\textbackslash n\textbackslash nAnd looked down one as far as I could\textbackslash n\textbackslash nTo where it bent in the undergrowth;\textbackslash n\textbackslash nThen took the other, as just as fair,\textbackslash n\textbackslash nAnd having perhaps the better claim,\textbackslash n\textbackslash nBecause it was grassy and wanted wear \\
\hline
\end{tabular}

\end{center}

\caption{Examples of our manually collected data. The prompt (prefix) is colored in brown. The numbers are the Accuracy (Eq. \ref{eq: acc}) of GPT2-XL on memorizing the sequences, where $100\%$ Accuracy means the true suffix can be fully reconstructed with greedy decoding.}
\label{tab:app_exampels}
\end{table*}

%% file: latex/Tables/app_examples2.tex
\begin{table*}[!t]
\begin{center}
\begin{tabular}{|p{2.5cm}|p{0.8cm}|p{11cm}|}
\hline
Mike Wall Bio & 100\% & \textcolor{brown}{\hspace{1 mm}Wall\textbackslash n\textbackslash nMichael was a science writer for the Idaho National Laboratory and has been an intern at Wired.com, The Salinas Californian newspaper,} and the SLAC National Accelerator Laboratory. He has also worked as a herpetologist and wildlife biologist. He has a Ph.D. in evolutionary biology from the University of Sydney, Australia, a bachelor's degree from the \\
\hline
Hardware & 100\% & \textcolor{brown}{PCs) may be defined as a desktop, floor standing, or portable microcomputer that includes a system unit having a central processing unit (CPU) and associated volatile} and non-volatile memory, including random access memory (RAM) and basic input/output system read only memory (BIOS ROM), a system monitor, a keyboard, one or more flexible diskette drives, a CD-ROM drive, \\
\hline
Contact Info of Skyhorse Publishing & 100\% & \textcolor{brown}{, or educational purposes. Special editions can also be created to specifications. For details, contact the Special Sales Department, Arcade Publishing, 307 West 36th Street}, 11th Floor, New York, NY 10018 or arcade@skyhorsepublishing.com.\textbackslash n\textbackslash nArcade Publishing® is a registered trademark of Skyhorse Publishing, Inc.®, a Delaware corporation.\textbackslash n\textbackslash nVisit\\
\hline
Meme & \phantom{1}98\% & \textcolor{brown}{\hspace{1 mm} a lot; that's great! It's a little awkward to ask, but we need your help. If you have already donated, we sincerely thank you.} We're not salespeople, but we depend on donations averaging \$14.76 and fewer than 1\% of readers give. If you donate just \$5.00, the price of your coffee, Catholic Online School could keep thriving. Thank \\
\hline
Malik Report & 100\% & \textcolor{brown}{\hspace{1 mm} check that allowed Dvorak to flick the puck over his shoulder…\textbackslash n\textbackslash nAbout The Malik Report\textbackslash n\textbackslash nThe Malik Report is a destination for all} things Red Wings-related. I offer biased, perhaps unprofessional-at-times and verbose coverage of my favorite team, their prospects and developmental affiliates. I've joined the Kukla's Korner family with five years of blogging under \\
\hline
Porn & 100\% & \textcolor{brown}{\hspace{1 mm} make love to her. She returned the favor with an amazing blowjob and a masterful fuck session...\textbackslash nENJOY!!!!\textbackslash n\textbackslash nThis entire website has} a voluntary content rating to block access by minors. This rating is compatible with microsoft internet explorer's content filtering function and\textbackslash nfacilitates website blocking software. For a tutorial on blocking this site click here.\textbackslash nCopyright bangbros. \\
\hline
Pokémon Fans & 100\% & \textcolor{brown}{\hspace{1 mm} We’re a group of Pokémon fans dedicated to providing the best place on the Internet for discussing ideas and sharing fan-made content. Welcome! We’re} glad you’re here.\textbackslash n\textbackslash nIn order to join our community we need you to create an account with us. Doing so will allow you to make posts, submit and view fan art and fan fiction, download fan-made games,\\
\hline
\end{tabular}

\end{center}

\caption{Examples of memorized sequences we collect from the Pile-dedupe. The prompt (prefix) is colored in brown. The numbers are the Accuracy (Eq. \ref{eq: acc}) of Pythia on memorizing the sequences, where $100\%$ Accuracy means the true suffix can be fully reconstructed with greedy decoding.}
\label{tab:app_exampels2}
\end{table*}

%% file: latex/Figs/appendix_heatmap_all.tex
\begin{figure*}[th]
  \centering
  \includegraphics[width=1.0\linewidth]{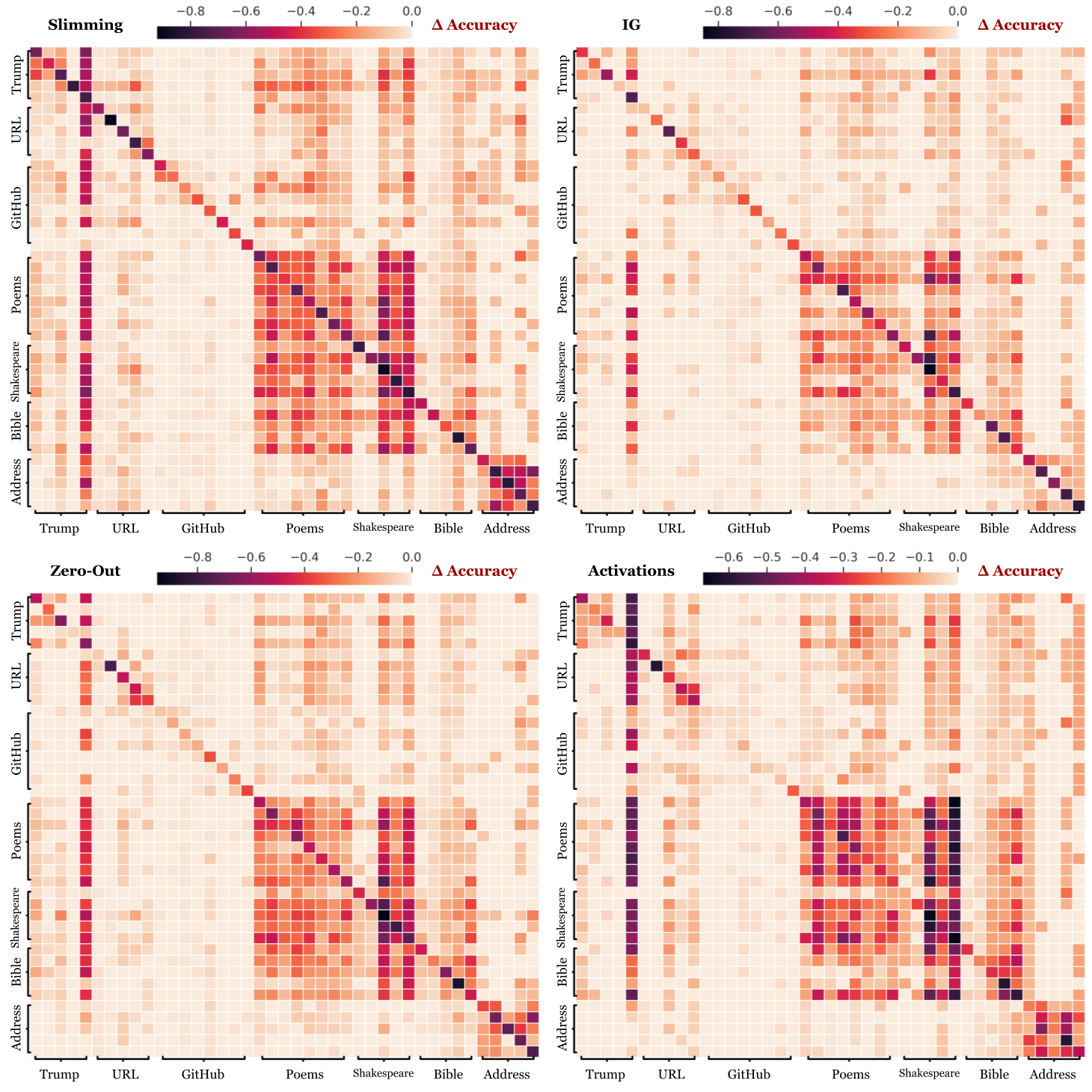}
  \caption{Confusion matrices of localization methods on a subset of sequences memorized by GPT2-XL, where each row/column represents a sequence. Different methods show similar patterns of confusion.}
  \label{fig:app_heatmap_all}
\end{figure*}

%% file: latex/Figs/appendix_heatmap_HC.tex
\begin{figure*}[th]
  \centering
  \includegraphics[width=1.0\linewidth]{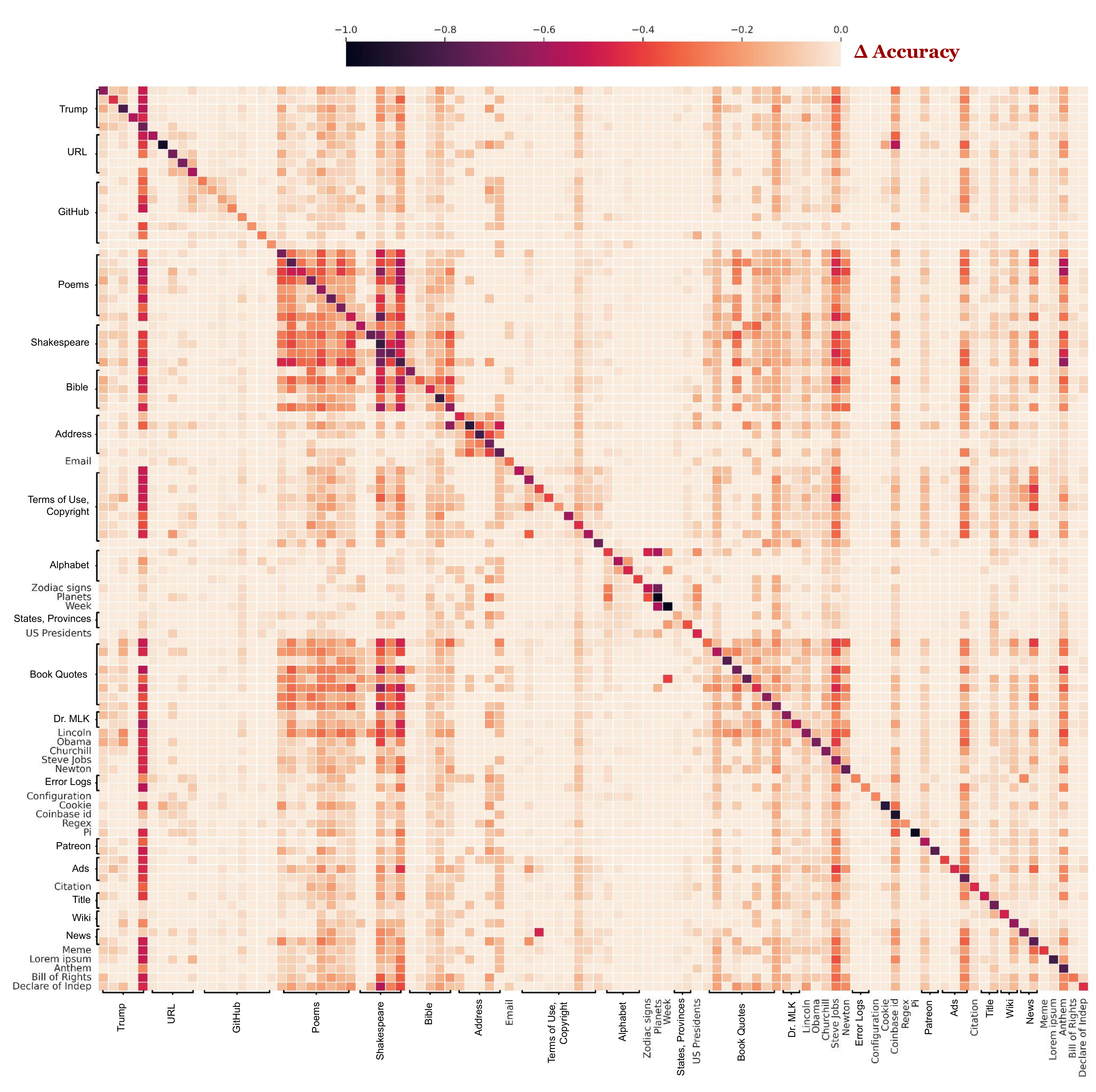}
  \caption{Confusion matrix of \HC\ on the entire test set memorized by GPT2-XL.
  Each row shows how dropping out the predicted neurons (0.5\%) on a target sequence changes the Accuracy of all sequences.
  \HC\ is unable to disentangle neurons of different quotes, including poems, Bible, books, and some famous people quotes.  Also, it finds a small set of neurons responsible for memorizing both the order of Zodiac Signs and the order of Planets.}
  \label{fig:app_heatmap_hc}
\end{figure*}